\RequirePackage{snapshot}
\documentclass[10pt,twocolumn,letterpaper]{article}

\usepackage{iccv}
\usepackage{times}
\usepackage{epsfig}
\usepackage{graphicx}
\usepackage{amsmath}
\usepackage{amssymb}

\usepackage{caption}
\usepackage{booktabs}
\usepackage{xspace}
\usepackage{xstring}
\usepackage{xparse}
\usepackage{xcolor}
\usepackage{multirow}
\usepackage{adjustbox}
\usepackage{setspace}
\usepackage[accsupp]{axessibility}  

\usepackage[pagebackref=true,breaklinks=true,letterpaper=true,colorlinks,bookmarks=false]{hyperref}
\usepackage[capitalise]{cleveref}

\iccvfinalcopy 

\usepackage{tikz}
\usepackage{pgfplots}
\pgfplotsset{compat=newest}
\usepgfplotslibrary{groupplots}
\usetikzlibrary{matrix}

\definecolor{ourswoggs}{RGB}{230, 171, 9}
\definecolor{ours}{RGB}{217, 95, 1}
\definecolor{colmapspsg}{RGB}{117, 112, 179}
\definecolor{colmapsift}{RGB}{231, 41, 138}
\definecolor{relpose}{RGB}{102, 102, 102}
\definecolor{pixsfm}{RGB}{196, 166, 166}

\crefname{section}{Sec.}{Secs.}
\Crefname{section}{Section}{Sections}
\Crefname{table}{Table}{Tables}
\crefname{table}{Tab.}{Tabs.}

\definecolor{maroon}{HTML}{efa884}
\definecolor{dorange}{HTML}{3b2000}
\hypersetup{citecolor=orange}
\hypersetup{linkcolor=orange}
\hypersetup{linkcolor=orange}
\hypersetup{urlcolor=orange}

\makeatletter
\newcommand{\oneptsmaller}[1]{%
  \begingroup
  \fontsize{\dimexpr\f@size pt-1pt}{\f@baselineskip}\selectfont
  #1%
  \endgroup
}
\makeatother

\makeatletter
\renewcommand{\paragraph}{%
  \@startsection{paragraph}{4}%
  {\z@}{0.25em}{-1em}%
  {\normalfont\normalsize\bfseries}}
\makeatother

\newcommand{\method}[1]{\oneptsmaller{\textsf{#1}}\xspace}
\newcommand{\clearname}{{{PoseDiffusion}}}
\newcommand{\name}{\oneptsmaller{\textsf{PoseDiffusion}}\xspace}
\newcommand{\namewoggs}{\oneptsmaller{\textsf{PoseDiffusion} w/o GGS}\xspace}

\newcommand{\N}{\mathbb{N}}
\newcommand{\R}{\mathbb{R}}
\newcommand{\T}{\mathbf{t}}
\newcommand{\q}{\mathbf{q}}
\newcommand{\p}{\mathbf{p}}
\newcommand{\I}{\mathtt{I}}

\newcommand{\ATE}{\mathrm{ATE}}
\newcommand{\ARE}{\mathrm{ARE}}
\newcommand{\RTE}{\mathrm{RTA}}
\newcommand{\RRE}{\mathrm{RRA}}
\newcommand{\AUC}{\mathrm{AUC}}
\newcommand{\mAA}{\mathrm{mAA}}

\def\iccvPaperID{7463} 

\ificcvfinal\fi

\begin{document}

\title{PoseDiffusion: Solving Pose Estimation via Diffusion-aided Bundle Adjustment}

\author{
Jianyuan Wang$^{1,2}$ \\
{\tt\small jianyuan@robots.ox.ac.uk}
\and 
Christian Rupprecht$^{1}$ \\
{\tt\small chrisr@robots.ox.ac.uk}
\and 
David Novotny$^2$ \\
{\tt\small dnovotny@meta.com }
\and
\\
$^1$Visual Geometry Group, University of Oxford \hspace{3.5em} 
$^2$Meta AI 
}



\twocolumn[{%
\renewcommand\twocolumn[1][]{#1}%
\maketitle
\thispagestyle{empty}
\begin{center}
\centering
\captionsetup{type=figure}
\includegraphics[width=1.01\linewidth,trim={0, 0, 0, 0},clip]{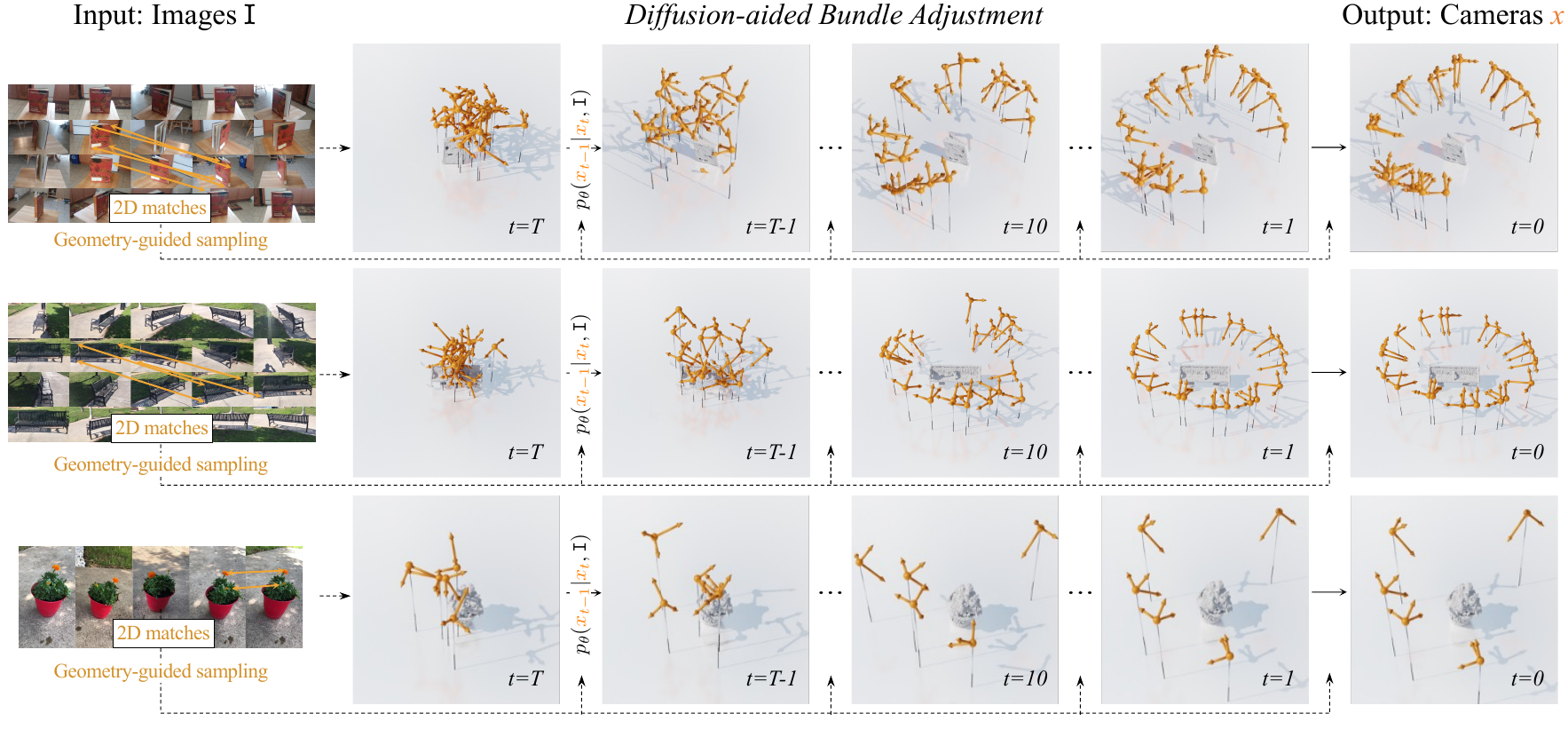}
\captionof{figure}{
\textbf{Camera Pose Estimation with \clearname.} We present a method to predict the camera parameters (extrinsics and intriniscs) for a given collection of scene images.
Our model combines the strengths of traditional epipolar constraints from point correspondences with the power of diffusion models to iteratively refine an initially random set of poses.
}%
\label{fig:teaser-fig}%
\end{center}%
\vspace{1em}%
}]

\begin{abstract}
Camera pose estimation is a long-standing computer vision problem that to date often relies on classical methods, such as handcrafted keypoint matching, RANSAC and bundle adjustment. 
In this paper, we propose to formulate the Structure from Motion (SfM) problem inside a probabilistic diffusion framework, modelling the conditional distribution of camera poses given input images.
This novel view of an old problem has several advantages. 
(i) The nature of the diffusion framework mirrors the iterative procedure of bundle adjustment.
(ii) The formulation allows a seamless integration of geometric constraints from epipolar geometry.
(iii) It excels in typically difficult scenarios such as sparse views with wide baselines.
(iv) The method can predict intrinsics and extrinsics for an arbitrary amount of images.
We demonstrate that our method \clearname \ significantly improves over the classic SfM pipelines and the learned approaches on two real-world datasets. 
Finally, it is observed that our method can generalize across datasets without further training. Project page: \url{https://posediffusion.github.io/}
\end{abstract}


\section{Introduction}


Camera pose estimation, \ie extracting the camera intrinsics and extrinsics given a set of free-form multi-view scene-centric images (\eg tourist photos of Rome \cite{agarwal_building_2009}), is a traditional Computer Vision problem with a history stretching long before the inception of modern computers \cite{kruppa1913ermittlung}.
It is a crucial task in various applications, including augmented and virtual reality, and has recently regained the attention of the research community due to the emergence of implicit novel-view synthesis methods \cite{mildenhall_nerf_2020,reizenstein_common_2021,lombardi_neural_2019}.

The classic dense pose estimation task estimates the parameters of many cameras with overlapping frusta, leveraging correspondence pairs between keypoints visible across images.
It is typically addressed through a Structure-from-Motion (SfM) framework, which not only estimates the camera pose (Motion) but also extracts the 3D shape of the observed scene (Structure). 
During the last 30 years, SfM pipelines matured into a technology capable of reconstructing thousands \cite{agarwal_building_2009} if not millions \cite{heinly_2015_reconstructing} of free-form views.

Surprisingly, the structure of dense-view SfM pipeline~\cite{schaffalitzky_multi-view_2002} has remained mostly unchanged until today, even though individual components have incorporated deep learning advances~\cite{detone2018self,sarlin2020superglue,jin2021image,tang2018ba, Wang_2021_CVPR, lindenberger2021pixel}.
SfM first estimates reliable image-to-image correspondences and, later, uses Bundle Adjustment (BA) to align all cameras into a common scene-consistent reference frame.
Due to the high complexity of the BA optimization landscape,
a modern SfM pipeline \cite{schonberger_structure--motion_2016} comprises a carefully engineered iterative process alternating between expanding the set of registered poses and a precise 2nd-order BA optimizer \cite{Agarwal_Ceres_Solver_2022}.

With the recent proliferation of deep geometry learning, the sparse pose problem, operating on a significantly smaller number of input views separated by wide baselines, has become of increasing interest.
For many years, this sparse setting has been the Achilles' Heel of traditional pose estimation methods. 
Recently, RelPose \cite{zhang2022relpose} leveraged a deep network to implicitly learn a bundle-adjustment prior from a large dataset of images and corresponding camera poses.
The method has demonstrated performance superior to SfM in settings with less than ten input frames.
However, in the many-image case, its accuracy cannot match the precise solution of the second-order BA optimizer from iterative SfM. Besides, it is limited to predicting rotations only.


In this paper, we propose \clearname \ - a novel camera pose estimation approach that elegantly marries deep learning with correspondence-based constraints and therefore, is able to reconstruct camera positions at high accuracy both in the sparse-view and dense-view regimes.



\clearname \ introduces a diffusion framework to solve the bundle adjustment problem by modelling the probability $p(x|\I)$ of camera parameters $x$ given observed images $\I$.
Following the recent successes of diffusion models in modelling complex distributions (\eg over images \cite{ho_denoising_2020}, videos \cite{singer2022make}, and point clouds \cite{luo_diffusion_2021}), we leverage diffusion models to learn $p(x|\I)$ from a large dataset of images with known camera poses.
Once learned, given a previously unseen sequence, we estimate the camera poses $x$ by sampling $p(x|\I)$.
The latter mildly assumes that $p(x|\I)$ forms a near-delta distribution so that any sample from $p(x|\I)$ will yield a valid pose. 
%
%
The stochastic sampling process of diffusion models has been shown to efficiently navigate the log-likelihood landscape of complex distributions \cite{ho_denoising_2020}, and therefore is a perfect fit for the intricate BA optimization.
An additional benefit of the diffusion process is that it can be trained one step at a time without the need for unrolling gradients through the whole optimization.


Additionally, in order to increase the precision of our camera estimation, we guide the sampling process with traditional epipolar constraints (expressed by means of reliable 2D image-to-image correspondences), which is inspired by classifier diffusion
guidance~\cite{dhariwal2021diffusion}.
We use this classical constraint to bias samples towards more geometrically consistent solutions throughout the sampling process, arriving at a more precise camera estimation.

%


\clearname \ yields state-of-the-art accuracy on the object-centric scenes of CO3Dv2 \cite{reizenstein_common_2021}, as well as on outdoor/indoor scenes of RealEstate10k \cite{zhou2018stereo}.
Crucially, \clearname \ also outperforms SfM methods when used to supervise NeRF training \cite{mildenhall_nerf_2020}, which demonstrates the superior accuracy of both the extrinsic and intrinsic estimation.


\section{Related Work}

\begin{figure*}[ht]
\centering
\includegraphics[width=\linewidth]{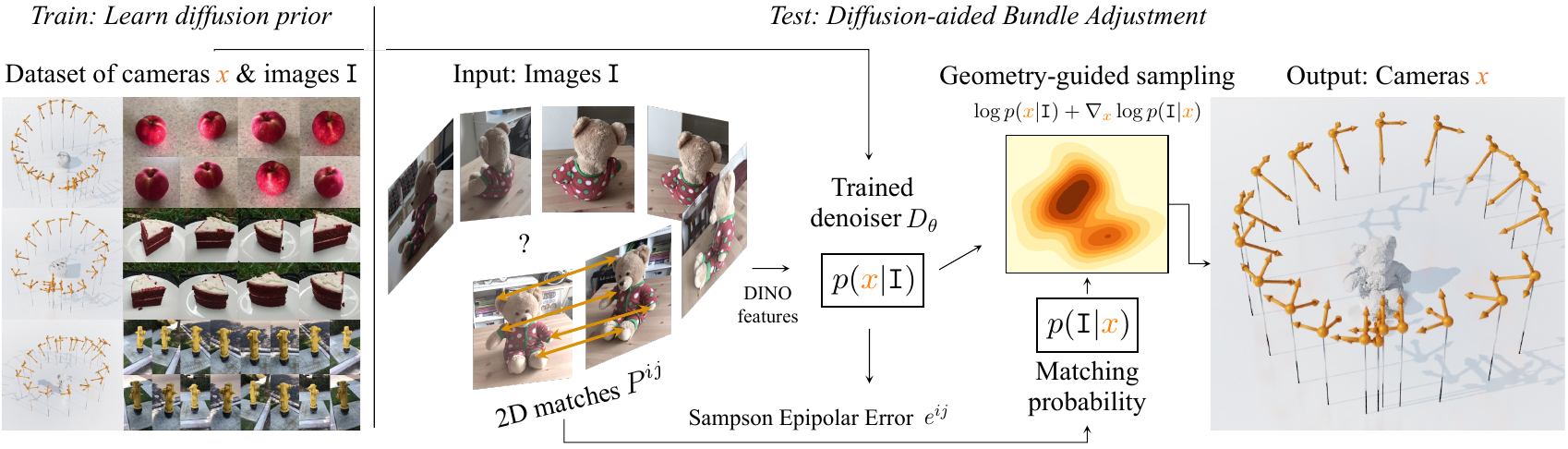}
\caption{\textbf{\clearname \ overview.} Training is supervised given a multi-view dataset of images and camera poses to learn a diffusion model $D_\theta$ to model $p(x|\I)$. During inference the reverse diffusion process is guided through the gradient that minimizes the Sampson Epipolar Error between image pairs, optimizing geometric consistency between poses.}
\label{fig:my_label}
\end{figure*}

\paragraph{Geometric Pose Estimation.}
The technique of estimating camera poses given image-to-image point correspondences has been extensively explored in the last three decades~\cite{hartley_multiple_2004, ozyecsil2017survey}.
%
%
This process typically begins with keypoint detection, conducted by handcrafted methods like SIFT~\cite{lowe_object_1999,lowe_distinctive_2004-1} and SURF~\cite{bay_speeded-up_2008}, or alternatively, learned methods~\cite{detone2018self, yi_lift_2016}.
%
The correspondences can then be established using nearest neighbour search or learned matchers~\cite{sarlin2020superglue,mao20223dg,zhang2019learning}.
Given these correspondences, five-point or eight-point algorithms compute camera poses~\cite{hartley_multiple_2004,hartley1997defense,li2006five,nister2004efficient} with the help of RANSAC and its variants~\cite{fischler_random_1981,brachmann2017dsac,brachmann2019neural}.
%
Typically, Bundle Adjustment~\cite{triggs_bundle_2000} further optimizes the camera poses.
The entire pipeline, from keypoint detection to bundle adjustment, is highly interdependent and needs careful tuning to be sufficiently robust, which allows for scaling to thousands of images~\cite{furukawa_towards_2010,sarlin2019coarse}.
{COLMAP} \cite{schonberger_structure--motion_2016,schoenberger2016mvs} is an open-source implementation of the whole camera estimation procedure and has become a valuable asset to the community.

\paragraph{Learned Pose Estimation.}
Geometric pose estimation techniques struggle when only few image-to-image matches can be established, or more generally, in a setting with sparse views and wide baselines~\cite{choi2015robust}. 
Thus, instead of constructing geometric constraints on top of potentially unreliable point matches, learning-based approaches directly estimate the camera motion between frames. 
Learning can be driven by ground truth annotations or unsupervisedly through reprojecting points from one frame to another, measuring photometric reconstruction~\cite{zhou2017unsupervised,ummenhofer2017demon,tang2018ba}.
Learned methods that directly predict the relative transformation between camera poses are often category-specific or object~centric~\cite{kehl2017ssd,xiang2017posecnn,ma2022virtual,wu2020unsupervised,wu2023magicpony}.
Recently, RelPose~\cite{zhang2022relpose} shows category-agnostic camera pose estimation, however, is limited to predicting rotations. 
The concurrent work SparsePose~\cite{sinha2023sparsepose} first regresses camera poses followed by iterative refinement, while RelPose++~\cite{lin2023relposepp} decouples the ambiguity in rotation estimation from translation prediction by defining a new coordinate system.



\paragraph{Diffusion Model.} Diffusion models are a category of generative models that, inspired by non-equilibrium thermodynamics~\cite{sohl2015deep}, approximate the data distribution by a Markov Chain of diffusion steps.
Recently, they have shown impressive results on image~\cite{song2019generative,ho_denoising_2020}, video~\cite{singer2022make,ho2022video}, and even 3D point cloud~\cite{luo_diffusion_2021, lyu2021conditional,Melas-Kyriazi_2023_CVPR} generation.
Their ability to accurately generate diverse high-quality samples has marked them as a promising tool in various fields.



\section{\clearname}

\paragraph{Problem setting.} We consider the problem of estimating intrinsic and extrinsic camera parameters given corresponding images of a single scene (\eg frames from an object-centric video, or free-form pictures of a scene).

Formally, given a tuple $\I = \big( I^i \big)_{i=1}^{N}$ of $N \in \N$ input images $I^i \in \R^{3\times H \times W}$, we seek to recover the tuple $x = \big( x^i \big)_{i=1}^N$ of corresponding camera parameters 
$x^i = (K^i, g^i)$ consisting of intrinsics
$K^i \subset \R^{3 \times 3}$
and extrinsics
$g^i \in \mathbb{SE}(3)$ respectively.
We defer the details of the camera parametrization to \cref{sec:method_details}.

Extrinsics $g^i$ map a 3D point $\p_w \in \R^3$ from world coordinates to a 3D point $\p_c \in \R^3 = g^i(\p_w)$ in camera coordinates.
Intrinsics $K^i$ then perspectivelly project $\p_c$ to a 2D point $\p_s \in \R^2$ in the screen coordinates with $K^i \p_c \sim \lambda [\p_s; 1], \lambda \in R$ where ``$\sim$'' indicates homogeneous equivalence.

\subsection{Preliminaries of Diffusion models} 
Diffusion models~\cite{ho_denoising_2020,sohl2015deep,song2019generative} are a class of likelihood-based models. They model a complex data distribution by learning to invert a diffusion process from data to a simple distribution, usually by means of noising and denoising.
The noising process gradually converts the data sample $x$ into noise  by a sequence of $T \in \N$ steps. The model is then trained to learn the \textit{de}noising process.
%

A Denoising Diffusion Probabilistic Model (DDPM) specifically defines the noising process to be Gaussian.
Given a variance schedule $\beta_1, ..., \beta_T$ of $T$ steps, the noising transitions are defined as follows:

\begin{equation}\label{eq:q}
q(x_t \mid x_{t-1}) = \mathcal{N}(x_t; \sqrt{1 - \beta_t} x_{t-1}, \beta_t \mathbb{I}),
\end{equation}
where $\mathbb{I}$ is the identity matrix.
The variance schedule is set so that $x_T$ follows an isotropic Gaussian distribution, \ie, $q(x_T) \approx \mathcal{N}(\mathbf{0}, \mathbb{I})$.
Define $\alpha_t = 1 - \beta_t$ and $\bar{\alpha}_t = \prod_{i=1}^t \alpha_i$, then a closed-form solution~\cite{ho_denoising_2020} exists to directly sample $x_t$ given a datum $x_0$: 
\begin{equation}
x_t \sim q(x_t \mid x_0) = \mathcal{N}(x_t; \sqrt{\bar{\alpha}_t} x_0, (1 - \bar{\alpha}_t) \mathbb{I}).
\end{equation}

The reverse $p_\theta(x_{t-1} | x_t)$ is still Gaussian if $\beta_t$ is small enough.
Therefore, it can be approximated by a model $\mathcal{D}_\theta$:
\begin{equation}\label{eq:p}
p_\theta(x_{t-1} \mid x_t) = \mathcal{N}(x_{t-1}; \sqrt{\alpha_t} \mathcal{D}_\theta(x_t, t), (1 - \alpha_t) \mathbb{I}).
\end{equation}



\subsection{Diffusion-aided Bundle Adjustment} 
\label{seq:pose_via_diffusion}
\clearname \ models the conditional probability distribution $p(x|\I)$ of the samples $x$ (\ie camera parameters) given the images $\I$.
Following the diffusion framework \cite{sohl2015deep} (discussed above), we model $p(x|\I)$ by means of the denoising process.
More specifically, $p(x|\I)$ is first estimated by training a diffusion model $\mathcal{D}_\theta$ on a large training set $\mathcal{T} = \{(x_j, \I_j)\}_{j=1}^S$ of $S \in \N$ scenes with ground truth image batches $\I_j$ and their camera parameters $x_j$.
At inference time, for a new set of observed images $\I$, we sample $p(x|\I)$ in order to estimate the corresponding camera parameters $x$. 
Note that, unlike for the noising process (\cref{eq:q}) which is independent of $\I$, the denoising process is conditioned on the input image set $\I$, \ie, $p_\theta(x_{t-1} \mid x_t, \I)$:
\begin{equation}\label{eq:p}
p_\theta(x_{t-1} | x_t, \I) = \mathcal{N}(x_{t-1}; \sqrt{\alpha_t} \mathcal{D}_\theta(x_t, t, \I), (1 - \alpha_t) \mathbb{I}).
\end{equation}

\paragraph{Denoiser $\mathcal{D}_\theta$.}
We implement the denoiser $\mathcal{D}_\theta$ as a Transformer $\mathrm{Trans}$ \cite{vaswani_attention_2017}:
\begin{equation}\label{eq:denoiser_transformer}
\mathcal{D}_\theta(x_t, t, \I) =  \mathrm{Trans}\left[\left(\text{cat}(x_t^i, t, \psi(I^i)\right)_{i=1}^N \right] = \mu_{t-1}.
\end{equation}
Here, $\mathrm{Trans}$ accepts a sequence of noisy pose tuples
$x_t^i$,
diffusion time $t$,
and feature embeddings $\psi(I^i) \in \R^{D_\psi}$ of the input images $I^i$.
The denoiser outputs the tuple of corresponding denoised camera parameters $\mu_{t-1} = (\mu_{t-1}^i)_{i=1}^N$.
Feature embeddings come from a vision transformer model initialized with weights of pre-trained DINO~\cite{caron2021dino}.


At train time, $\mathcal{D}_\theta$ is supervised with the denoising loss:
\begin{equation}\label{eq:p_superv}
\mathcal{L}_\text{diff} = E_{t \sim [1, T], x_t \sim q(x_t | x_0, \I)} \|
    \mathcal{D}_\theta(x_t, t, \I) - x_0
\|^2,
\end{equation}
where the expectation aggregates over all diffusion time-steps $t$, the corresponding diffused samples $x_t \sim q(x_t | x_0, \I)$, and a training set $\mathcal{T} = \{(x_{0,j}, \I_j)\}_{j=1}^S$ of $S \in \N$ scenes with images $\I_j$ and their cameras $x_{0,j}$.

\paragraph{Solving Bundle Adjustment by Sampling $p_\theta$.} \label{sec:ddpm_sampling}
The trained denoiser
$\mathcal{D}_\theta$
(\cref{eq:p_superv}) is later leveraged to sample
$p_\theta(x|\I)$
which effectively solves our task of inferring camera parameters $x$ given input images $\I$.
Note that we assume $p(x|\I)$ forms a near-delta distribution and, hence, any sample from $p(x|\I)$ will yield a valid pose.
Such mild assumption allows to avoid a maximum-aposteriori probability (MAP) estimate of $p(x|\I)$.

In more detail, following DDPM sampling \cite{ho_denoising_2020}, we start from random cameras
$x_T \sim \mathcal{N}(\mathbf{0}, \mathbb{I})$ and, in each iteration $t \in (T, ..., 0)$, the next step $x_{t-1}$ is sampled from
\begin{equation}\label{eq:ddpm_sampling}
x_{t-1} \sim \mathcal{N}(
x_{t-1} ;
\sqrt{\bar{\alpha}_{t-1}} \mathcal{D}_\theta(x_t, t, \I), (1-\bar{\alpha}_{t-1}) \mathbb{I}
)
.
\end{equation} 

\begin{figure}[t] \label{fig:samplingstep}
\centering
\includegraphics[width=1.01\linewidth]{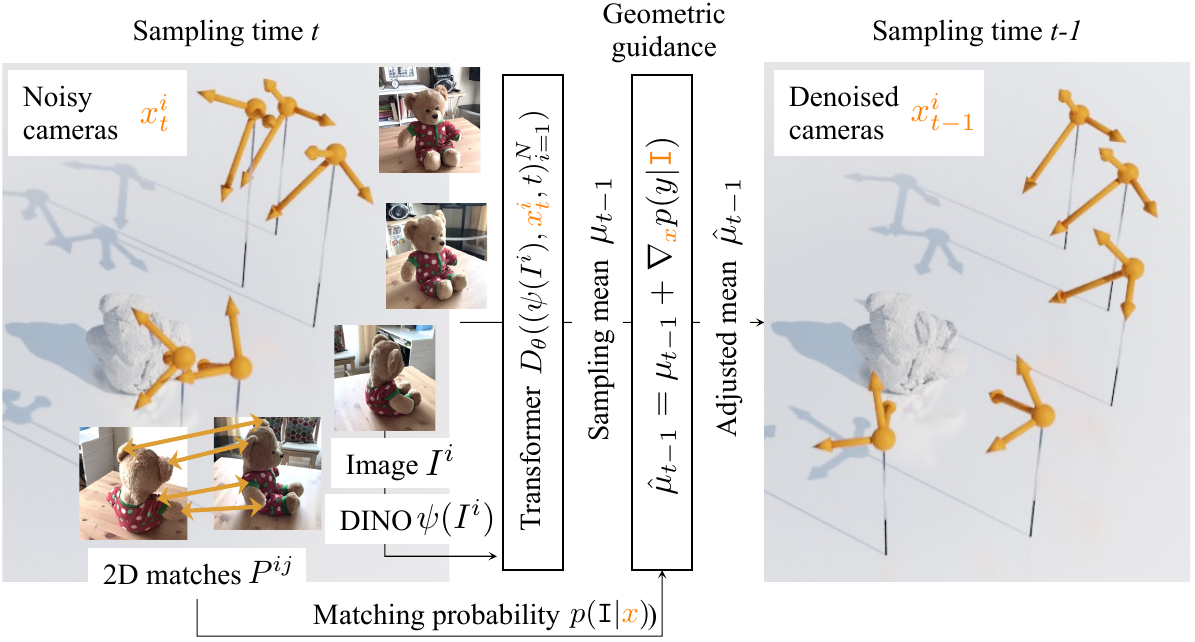}
\caption{%
\textbf{Inference.} For each step $t$, Geometry-Guided Sampling (GGS)
samples the distribution $p_\theta(x_{t-1} \mid x_t, \I, t)$
of refined cameras $x_{t-1}$ conditioned on input images $\I$ and the previous estimate $x_t$, while being guided by the gradient of the Sampson matching density $p(\I|x)$.
}
\end{figure}

\subsection{Geometry-Guided sampling}
So far, our feed-forward network maps images directly to the space of camera parameters.
Since deep networks are notoriously bad at regressing precise quantities, such as camera translation vectors or angles of rotation matrices \cite{kendall2015posenet}, we significantly increase the accuracy of \clearname \ by leveraging two-view geometry constraints which form the backbone of state-of-the-art SfM methods.

To this end, we extract reliable 2D correspondences between scene images and guide DDPM sampling iterations (\cref{eq:ddpm_sampling}) so that the estimated poses satisfy the correspondence-induced two-view epipolar constraints.

\paragraph{Sampson Epipolar Error.} Specifically, let $P^{ij} = \{ (\p_k^i, \p_k^j) \}_{k=1}^{N_{P^{ij}}}$ denote a set of 2D correspondences between image points $\p_k \in \R^2$ for a pair of scene images $(I^i, I^j)$, and denote $(x^i, x^j)$ the corresponding camera poses.
Given the latter, we evaluate the compatibility between the cameras and the 2D correspondences via a robust version of Sampson Epipolar Error $e^{ij} \in \R$ \cite{hartley_multiple_2004}:
\begin{equation*}
\label{eq:sampson}
\begin{gathered}
e^{ij}(x^i, x^j, P^{ij}) = \\
\sum_{k=1}^{|P^{ij}|} 
\left[
\frac{
    {\tilde{\p}_k^{j \top}} F^{ij} {\tilde{\p}_k^i}
}{
    (F^{ij} {\tilde{\p}_k^i})^2_1 + (F^{ij} {\tilde{\p}_k^i})^2_2
    + ({F^{ij \top}} {\tilde{\p}_k^j})^2_1 + ({F^{ij \top}} {\tilde{\p}_k^j})^2_2
}
\right]_\epsilon,
\end{gathered}
\end{equation*}
where $\tilde{\p} = [\p; 1]$ denotes $\p$ in homogeneous coordinates,
$\left[z\right]_\epsilon = \min(z, \epsilon)$ is a robust clamping function,
$(\mathbf{z})_l$ retrieves $l$-th element of a vector $\mathbf{z}$,
and $F^{ij} \in \R^{3 \times 3}$ is the Fundamental Matrix \cite{hartley_multiple_2004} mapping points $\p_k^i$ from image $I^i$ to lines in image $I^j$ and vice-versa. Directly optimizing the epipolar constraint ${\tilde{\p}_k^{j \top}} F^{ij} {\tilde{\p}_k^i}$ usually provides sub-optimal results~\cite{hartley_multiple_2004}, which is also observed in our experiments. 



\graphicspath{{figures/qualitative/co3dv2}}

\newcommand{\figwidth}{2.45cm}%
\newcommand{\figheight}{2.0cm}%

\newcommand{\figwidthim}{2.3cm}%
\newcommand{\figheightim}{2.0cm}%

\newcommand{\cothreeddqfig}[6]{%
\adjustbox{trim={#6\width} {#6\height} {#6\width} {#6\height},clip,width=\figwidth}%
{\includegraphics[]{{#1_nf#2_#3_#4_vs_gt_#5_000.png}}}%
}

\newcommand{\imcol}[5]{%
\begin{tabular}{c}%
\begin{minipage}[c][\figheight][c]{\figwidth}%
\centering%
\includegraphics[height=\figheightim,width=\figwidthim,keepaspectratio]{{#1_nf#2_#3_imgrid.png}}%
\end{minipage}\vspace{0.1cm}\\%
\cothreeddqfig{#1}{#2}{#3}{ours}{#4}{#5}\\%
\cothreeddqfig{#1}{#2}{#3}{relpose}{#4}{#5}\\%
\cothreeddqfig{#1}{#2}{#3}{colmapspsg}{#4}{#5}\\%
\end{tabular}
}

\newcommand{\imcolone}[5]{%
\small\centering%
\begin{tabular}{ccc}%
\rotatebox{90}{\hspace{-0.2cm}Input $\I$}&\phantom{a}&%
\begin{minipage}[c][\figheight][c]{\figwidth}%
\centering\includegraphics[height=\figheightim,width=\figwidthim,keepaspectratio]{{#1_nf#2_#3_imgrid.png}}%
\end{minipage}\vspace{0.1cm}\\%
\rotatebox{90}{\hspace{0.25cm}\textbf{\name}}&&\cothreeddqfig{#1}{#2}{#3}{ours}{#4}{#5}\\%
\rotatebox{90}{\hspace{0.75cm}\method{RelPose}}&&\cothreeddqfig{#1}{#2}{#3}{relpose}{#4}{#5}\\%
\rotatebox{90}{\hspace{0.2cm}\method{COLMAP+SPSG}}&&\cothreeddqfig{#1}{#2}{#3}{colmapspsg}{#4}{#5}%
\end{tabular}
}

\begin{figure*}
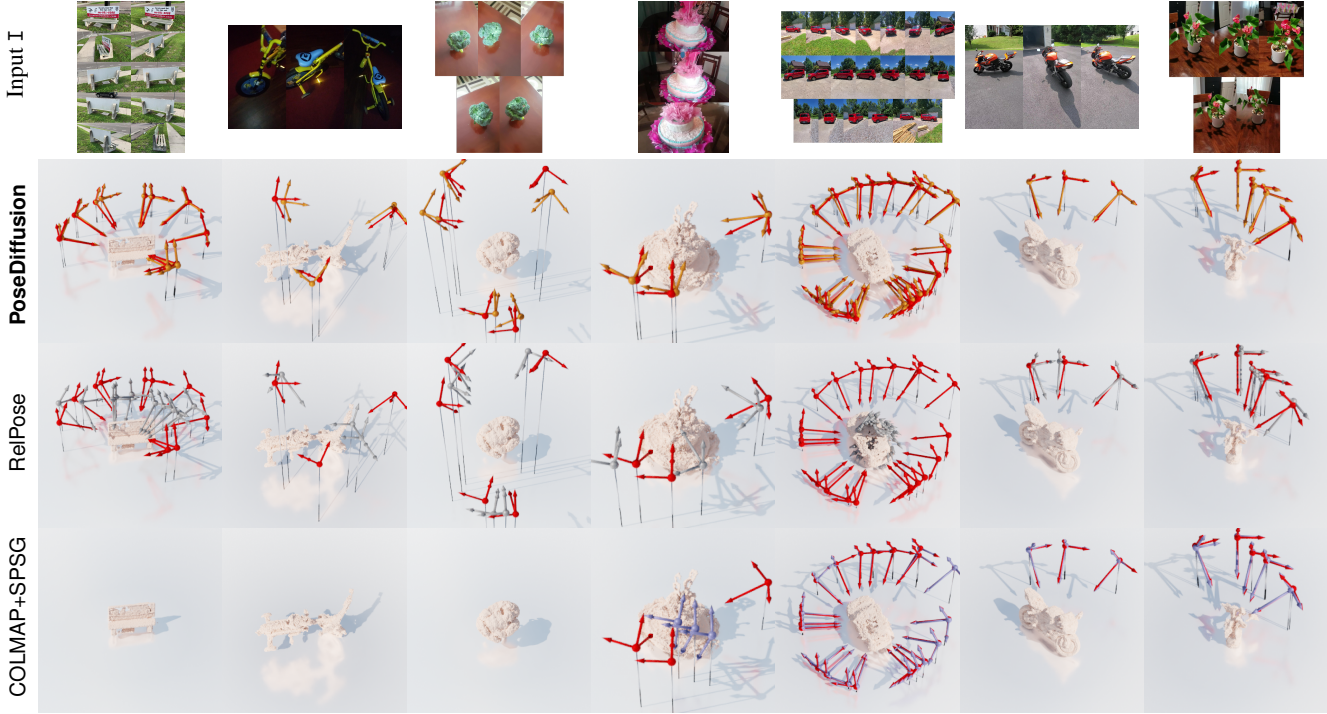

\setlength\tabcolsep{0pt}%
\renewcommand{\arraystretch}{0.0}%
\imcolone{bench}{10}{0000}{000}{0.15}%
\imcol{bicycle}{3}{0001}{045}{0.1}%
\imcol{broccoli}{5}{0000}{000}{0.1}%
\imcol{cake}{3}{0001}{000}{0.22}%
\imcol{car}{20}{0001}{135}{0.15}%
\imcol{motorcycle}{3}{0000}{135}{0.17}%
\imcol{plant}{5}{0000}{135}{0.2}%
\caption{\textbf{Pose estimation on CO3Dv2.} Estimated cameras given input images $\I$ (first row).
Our {\color{ours}\name}\ (2nd row) is compared to {\color{relpose}\method{RelPose}} (3rd row), {\color{colmapspsg}\method{COLMAP+SPSG}} (4th row), and the {\color{red} ground truth}.
Missing cameras indicate failure. 
\label{fig:co3d_figs}%
}
\end{figure*}

\paragraph{Sampson-guided sampling.}
We follow the classifier diffusion guidance \cite{dhariwal2021diffusion} to guide the sampling towards a solution which minimizes the Sampson Epipolar Error and, as such, satisfies the image-to-image epipolar constraint.

In each sampling iteration, classifier guidance perturbs the predicted mean $\mu_{t-1} = \mathcal{D}_\theta(x_t, t, \I)$ with a gradient of $x_t$-conditioned guidance distribution $p(\I | x_t)$:
\begin{equation}\label{eq:D_classifier}
\hat{\mathcal{D}}_\theta(x_t, t, \I) = \mathcal{D}_\theta(x_t, t, \I) + s \nabla_{x_t} \log p(\I | x_t),
\end{equation}
where $s \in \R$ controls the strength of the guidance.
$\hat{\mathcal{D}}_\theta(x_t, t, \I)$ then replaces $\mathcal{D}_\theta(x_t, t, \I)$ in \cref{eq:p,eq:ddpm_sampling}.

Assuming a uniform prior over cameras $x$ allows modeling $p(\I | x_t)$ from \cref{eq:D_classifier} as a product of independent exponential distributions over the pairwise Sampson Errors $e^{ij}$:
\begin{equation}\label{eq:sampson_classifier}
p(\I | x_t) = \prod_{i,j} p(I^i, I^j | x_t^i, x_t^j) \propto \prod_{i,j} \exp(- e^{ij}).  
\end{equation}
Note that our choice of $p(\I | x_t)$ is meaningful since its mode is attained when Sampson Errors between all image pairs is 0 (\ie all epipolar constraints are satisfied).


\subsection{Method details} \label{sec:method_details}

\paragraph{Representation details.} 
We represent the extrinsics $g^i = (\q^i, \T^i)$ as a 2-tuple comprising the quaternion $\q^i \in \mathbb{H}$ of the rotation matrix $R^i \in \mathbb{SO}(3)$ and the camera translation vector $\T^i \in \R^3$.
As such, $g^i(\p_w)$ represents a linear world-to-camera transformation $\p_c = g^i(\p_w) = R^i \p_w + \T^i$.
We use a camera calibration matrix
$K^i = \left[f^i, 0, p_x; 0, f^i, p_y; 0, 0, 1 \right] \in \R^{3\times3}$,
with one degree of freedom defined by the focal length $f^i \in \R^+$.
Following common practice in SfM~\cite{schoenberger2016sfm,schoenberger2016mvs}, the principal point coordinates $p_x, p_y \in \R$ are fixed to the center of the image.
To ensure strictly positive focal length $f^i$, we represent it as $f^i = \exp(\hat{f}^i)$, where $\hat{f}^i \in \R$ is the quantity predicted by the denoiser $\mathcal{D}_\theta$.
Therefore, the transformer $\mathrm{Trans}$ (\cref{eq:denoiser_transformer}) outputs a tuple of raw predictions $\left( (\hat{f}^i, \q^i, \T^i) \right)_{i=1}^N$ which is converted (in close-form) to a tuple of cameras $x = \left( (K^i, g^i) \right)_{i=1}^N$.

\paragraph{Tackling Coordinate Frame Ambiguity.} Because our training set $\mathcal{T}$ is constructed by SfM reconstructions~\cite{schoenberger2016sfm}, the training poses $\{\hat{g}_j\}_{j=1}^S$ are defined up to an arbitrary scene-specific similarity transformation.
To prevent overfitting to the scene-specific training coordinate frames, we canonicalize the input before passing to the denoiser:
we normalize the extrinsics $g_j = (\hat{g}^1_j, ... \hat{g}^N_j)$ as relative camera poses to a randomly selected pivot camera $\hat{g}_j^\star$.
We inform the denoiser about the pivot camera by appending a binary flag $p_\text{pivot}^i \in \{0, 1\}$ to the image features $\psi(I^i)$ (\cref{eq:denoiser_transformer}). 
Furthermore, in order to canonicalize the scale, we divide the input camera translations by the median of the norms of the pivot-normalized translations. 

\begin{figure*}[h]
\vspace{-0.2cm}%
\includegraphics[width=1.0\linewidth]{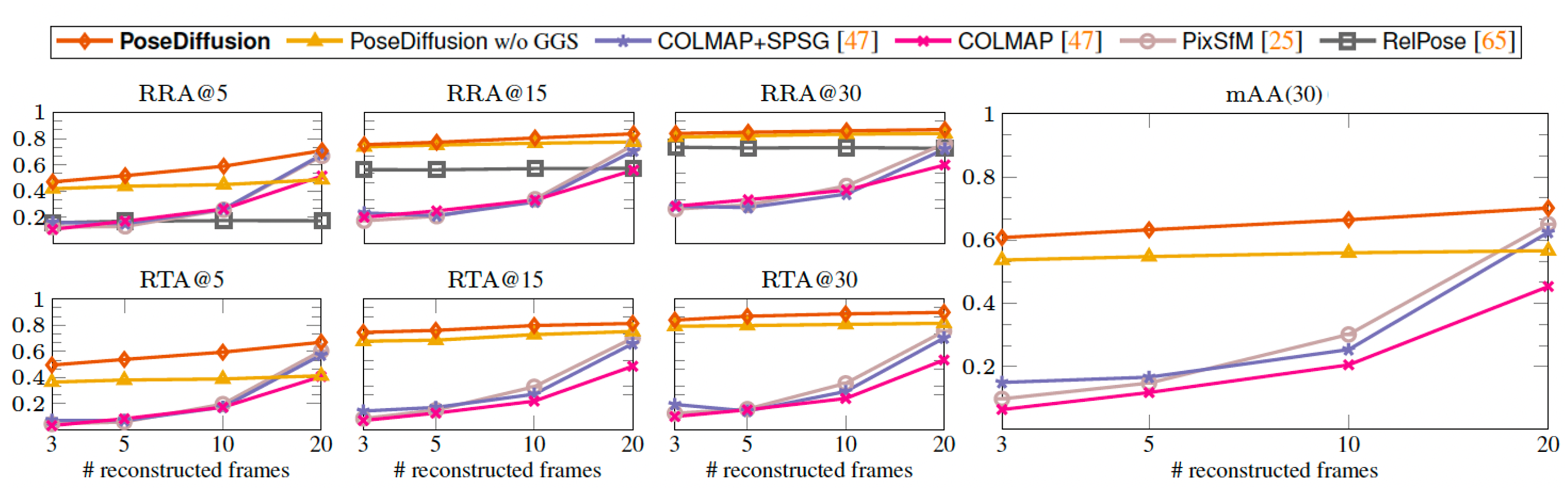}
\caption{
\label{fig:co3d_plots}
\textbf{Pose estimation accuracy on CO3Dv2.} Metrics $\RRE@\tau,\RTE@\tau$ at different thresholds $\tau$ and $\mAA(30)$ ($y$-axes, higher-better) as a function of the number of input frames ($x$-axes).
\method{RelPose} does not predict camera translation and hence is omitted in the respective figures. 
}%

\end{figure*}

\section{Experiments}
We experiment on two real-world datasets, ablate the design choices of the model, and compare with prior work.

\paragraph{Datasets.}
We consider two datasets with different statistics.
The first is \textbf{CO3Dv2} \cite{reizenstein_common_2021} containing roughly 37k turn-table-like videos of objects from 51 MS-COCO categories \cite{lin_microsoft_2014}.
The dataset provides cameras automatically annotated by COLMAP \cite{schonberger_structure--motion_2016} using 200 frames in each video. 
Secondly, we evaluate on \textbf{RealEstate10k} \cite{zhou2018stereo} which comprises 80k YouTube clips capturing the interior and exterior of real estate.
Its camera annotations were auto-generated with ORB-SLAM 2 \cite{mur-artal_orb-slam_2015} and refined with bundle adjustment.
We use the same training set as in \cite{wiles_synsin_2020}, \ie 57k training scenes and, as some baselines are time-consuming, a random smaller 1.8k-video subset of the original 7K test videos.
Naturally, we always test on unseen videos.

\paragraph{Baselines and comparisons.}
We chose \method{COLMAP} \cite{schonberger_structure--motion_2016}, one of the most popular SfM pipelines, as a dense-pose estimation baseline.
Besides the classic version leveraging RANSAC-matched SIFT features, we also benchmark \method{COLMAP+SPSG} which builds on SuperPoints \cite{detone2018self} matched with SuperGlue \cite{sarlin2020superglue}. 
\method{PixSfM}~\cite{lindenberger2021pixel} further improves accuracy by directly aligning deep features.
We also compare to \method{RelPose} \cite{zhang2022relpose} which is the current State of the Art in sparse pose estimation.
Finally, to ablate Geometry Guided Sampling (GGS - \cref{eq:sampson_classifier}), \namewoggs leverages our denoiser without GGS.

\paragraph{Training.}
We train the denoiser $\mathcal{D}_\theta$ using the Adam optimizer with the initial learning rate of 0.0005 until convergence of $\mathcal{L}_\text{diff}$ - learning rate is decayed ten-fold after 30 epochs.
The latter takes two days on 8 GPUs.
In each training batch, we randomly sample between 3-20 frames and their cameras from a random scene of the training dataset. 

\paragraph{Geometry-guided sampling.} 
\name's GGS leverages the SuperPoint features \cite{detone2018self} matched with SuperGlue \cite{sarlin2020superglue}, where the Sampson error is clamped at $\epsilon=10$ (\cref{eq:sampson}).
To avoid spurious local minima, we apply GGS to the last 10 diffusion sampling steps.
During each step $t$, we adjust the sampling mean by running 100 GGS iterations.
We observed improved sampling stability when the guidance strength
$s$ (\cref{eq:D_classifier}) is set adaptively 
so that the norm of the guidance gradient $\nabla p(\I | x)$ does not exceed a multiple $\alpha \| \mu_t \|$ ($\alpha=0.0001$) of the current mean's norm.



\paragraph{Evaluation metrics.}
We compute the \textbf{Relative Rotation Accuracy} ($\RRE$)
to compare the relative rotation $R_i R_j^\top$ from $i$-th to $j$-th camera to the ground truth $R_i^\star R_j^{\star \top}$.
Similarly, the \textbf{Relative Translation Accuracy}
$\mathbf{\RTE}(\T_{ij}, \T_{ij}^\star) = \arccos(\T_{ij}^\top \T_{ij}^\star / (\| \T_{ij} \| \| \T_{ij}^\star \|) )$
evaluates the angle between the predicted and ground-truth vector $\T_{ij}$ / $\T_{ij}^\star$ pointing from camera $i$ to $j$.
$\RRE/\RTE$ are invariant to the absolute coordinate frame ambiguity. 
For a given threshold $\tau$, we report  $\RTE@\tau/\RRE@\tau$ ($\tau \in \{5, 15, 30\}$), \ie the percentage of camera pairs with $\RRE/\RTE$ below a threshold $\tau$.

Additionally, following the Image Matching Benchmark~\cite{jin2021image}, we report \textbf{mean Average Accuracy} ($\mAA$) (also known as Area under Curve - $\AUC$).
Specifically, $\mAA$ calculates the area under the curve recording the accuracies of the angular differences between the ground-truth and predicted cameras for a range of angular accuracy thresholds.
For an image pair, $\mAA$ defines the accuracy at a threshold $\tau$ as $min(\RRE@\tau, \RTE@\tau)$.
Following RelPose's~\cite{zhang2022relpose} upper angular threshold of $30^{\circ}$, we report $\mAA(30)$ which is integrated over $\tau \in [1, 30]$.


%



\graphicspath{{figures/qualitative/re10k}}

\newcommand{\figwidthre}{2.7cm}%
\newcommand{\figheightre}{2.0cm}%

\newcommand{\refig}[6]{%
\adjustbox{trim={#6\width} {#6\height} {#6\width} {#6\height},clip,width=\figwidthre}%
{\includegraphics[]{{#1_nf#2_#3_#4_vs_gt_#5_000.png}}}%
}

\newcommand{\imcolre}[7]{%
\begin{tabular}{ccc}%
\multicolumn{3}{c}{
    \begin{minipage}[b][1.5cm][c]{6cm}%
    \centering\includegraphics[height=1.5cm,width=6cm,keepaspectratio]{{#1_nf#2_#3_imgrid.png}}%
    \end{minipage}%
}\\%
\refig{#1}{#2}{#3}{ours}{#4}{#7}&%
\refig{#1}{#2}{#3}{ours}{#5}{#7}&%
\refig{#1}{#2}{#3}{ours}{#6}{#7}\\%
\refig{#1}{#2}{#3}{colmapspsg}{#4}{#7}&%
\refig{#1}{#2}{#3}{colmapspsg}{#5}{#7}&%
\refig{#1}{#2}{#3}{colmapspsg}{#6}{#7}\\%
\end{tabular}%
}

\newcommand{\imcolonere}[7]{%
\small\centering%
\begin{tabular}{ccccc}%
\rotatebox{90}{\hspace{0.7cm}Input $\I$}%
&%
\phantom{a}&%
\multicolumn{3}{c}{
    \begin{minipage}[b][1.5cm][c]{6cm}%
    \centering\includegraphics[height=1.5cm,width=6cm,keepaspectratio]{{#1_nf#2_#3_imgrid.png}}%
    \end{minipage}%
}\\%
\rotatebox{90}{\hspace{0.5cm}\textbf{\name}}&&%
\refig{#1}{#2}{#3}{ours}{#4}{#7}&%
\refig{#1}{#2}{#3}{ours}{#5}{#7}&%
\refig{#1}{#2}{#3}{ours}{#6}{#7}\\%
\rotatebox{90}{\hspace{0.5cm}\method{COLMAP+SPSG}}&&
\refig{#1}{#2}{#3}{colmapspsg}{#4}{#7}&%
\refig{#1}{#2}{#3}{colmapspsg}{#5}{#7}&%
\refig{#1}{#2}{#3}{colmapspsg}{#6}{#7}\\%
\end{tabular}%
}

\begin{figure*}
\setlength\tabcolsep{0pt}%
\renewcommand{\arraystretch}{0.0}%
\imcolonere{realestate}{5}{0002}{000}{020}{040}{0.0}%
\imcolre{realestate}{10}{0008}{000}{020}{040}{0.0}%
\caption{\textbf{Pose estimation on RealEstate10k} visualizing the cameras estimated given input images $\I$ (first row).
Our {\color{ours}\name}\ (2nd row) is compared to {\color{colmapspsg}\method{COLMAP+SPSG}} (3rd row), and the {\color{red} ground truth}.
Missing cameras indicate failure. For  better visualization, we display each scene from three different viewpoints.
\label{fig:re10k_figs}%
}
\end{figure*}

\begin{figure*}[h]
\vspace{-0.2cm}%
\includegraphics[width=1.0\linewidth]{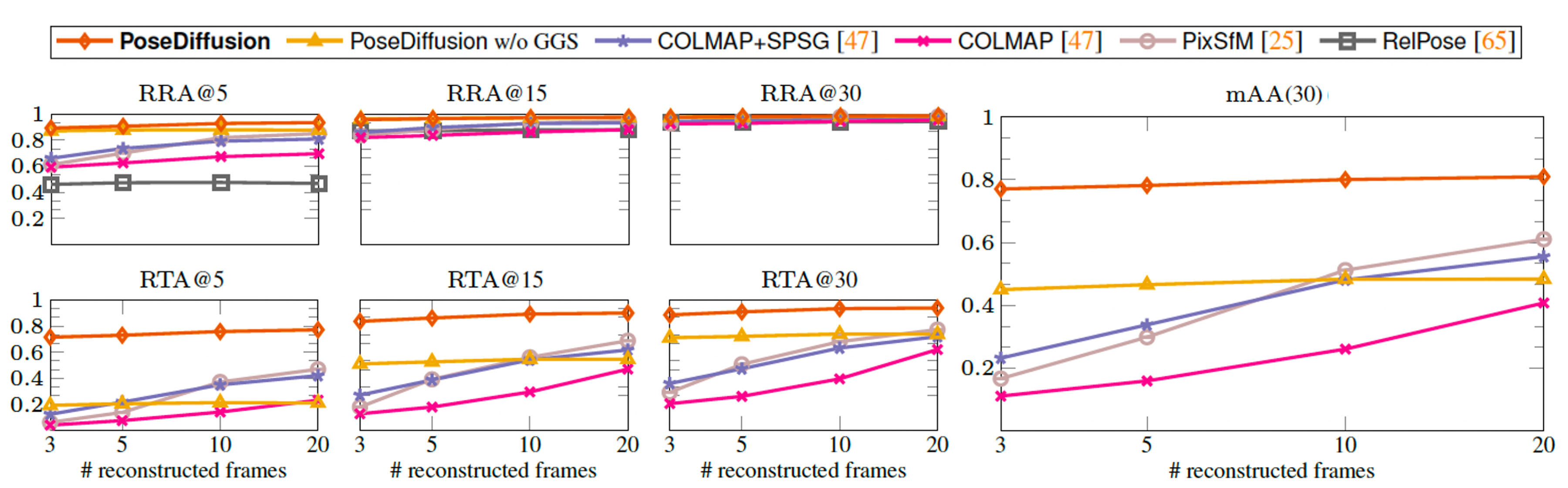}
\caption{
\textbf{Pose estimation on RealEstate10k.} Metrics $\RRE@\tau,\RTE@\tau$ at different thresholds $\tau$ and $\mAA(30)$ ($y$-axes, higher-better) as a function of the number of input frames ($x$-axes).
\label{fig:re10k_plots}
}%
\end{figure*}

\subsection{Camera pose estimation} \label{exp_pose_estimation}

 
\paragraph{Object-centric pose.}
We first compare on CO3Dv2 where each scene comprises frames capturing a single object from a variety of viewpoints with approximately constant distance from the object.
\cref{fig:co3d_plots} contains quantitative results while \cref{fig:co3d_figs} illustrates example camera estimates.
\name significantly improves over all baselines in all metrics in both the sparse and dense setting.
Note that, here ground truth cameras were obtained with \method{COLMAP} itself (but using 200 frames), likely favouring \method{COLMAP} reconstructions. 
Importantly, removing GGS (\namewoggs) leads to a drop in performance for tighter accuracy thresholds across all metrics.
This clearly demonstrates that GGS facilitates accurate camera estimates.
The latter also validates the accuracy of our intrinsics since they are an important component of GGS.

\paragraph{Scene-centric pose.}
Here, we reconstruct camera poses in free-form in/outdoor scenes of RealEstate10k which, historically, has been the domain of traditional SfM methods.
We evaluate quantitatively in \cref{fig:re10k_plots} and qualitatively in \cref{fig:re10k_figs}.
\name significantly outperforms all baselines in all metrics.
Here, the comparison to \method{COLMAP} is fairer than on CO3Dv2, as RealEstate10k used ORB-SLAM2~\cite{mur2015orb} to obtain the ground-truth cameras.


\paragraph{Importance of diffusion.} To validate the effect of the diffusion model, we also provide the \method{PoseReg} baseline, which uses the same architecture and training hyper-parameters as our method but directly regresses poses.
\method{PoseReg} shows clearly lower performance (cf.~\cref{tab:diffuse_or_nor}). Moreover, without the iterative refinement of our diffusion model, the gain of applying GGS to \method{PoseReg} (\method{PoseReg+GGS}) is limited.

\paragraph{Generalization.}
We also evaluate the ability of different methods to generalize to different data distributions.
First, following RelPose \cite{zhang2022relpose}, we train on a set of 41 training categories from CO3Dv2, and evaluate the remaining 10 held-out categories.
As shown in ~\cref{tab:generalization_categories_co3dv2}, our method outperforms all baselines indicating superior generalizability, even without the help of GGS.


Moreover, we evaluate a significantly more difficult scenario: transfer from the CO3Dv2 to RealEstate10k. 
This setting brings a considerable difficulty: CO3Dv2 predominantly contains indoor objects with circular fly-around trajectories while RealEstate10k comprises outdoor scenes and linear fly-through camera motion (see ~\cref{fig:co3d_figs,fig:re10k_figs}).
%
%
Surprisingly, our results are still comparable to  \method{PixSfM}, while better than  \method{COLMAP} and \method{RelPose}. 

\begin{table}[t]
{\centering\scriptsize
        \setlength\tabcolsep{2pt}%
        \begin{tabular}{c|ccccccc}
        \toprule
        \multirow{2}{*}{Metric} & \multirow{2}{*}{RelPose} & COLMAP  &  \multirow{2}{*}{PixSfM}   & \multirow{2}{*}{PoseReg}   & Ours  & PoseReg & \multirow{2}{*}{Ours} \\
        & & +SPSG  &  &  & w/o GGS & +GGS & \\
        \midrule 
        $\RRE@15$ & 57.1 & 31.6 & 33.7 & 53.2 &  \underline{75.9} & 57.0 & \textbf{80.5}   \\
        $\RTE@15$ & - & 27.3 & 32.9 & 49.1  & \underline{72.8} & 53.4 & \textbf{79.8} \\
        $\mAA(30)$ & - & 25.3 & 30.1 &  45.0  & \underline{56.0} & 48.2 & \textbf{66.5}     \\

        \bottomrule
        \end{tabular}
  \caption{
      \textbf{Pose regression ablation} comparing a diffusion-free pose regressor \method{PoseReg} (with/without GGS) to our \name on CO3Dv2 with 10 input frames (\textbf{Bold} denotes the top result and an \underline{underline} signifies the second best).
      \label{tab:diffuse_or_nor}
  }    
}
\end{table}

\begin{table}[t]
{\centering\scriptsize
        \setlength\tabcolsep{3pt}%
        \begin{tabular}{c|ccccc}
        \toprule
        {Test Set} & {COLMAP} & COLMAP+SPSG  & {PixSfM}   & Ours w/o GGS  & {Ours} \\
        \midrule
        {CO3Dv2 Unseen}  & 25.8 & 30.3 & 34.2 & 
 \underline{40.1} & \textbf{50.8}   \\
        {RealEstate10k}  & 26.1 &  45.2 & \textbf{49.4} & 18.7 & \underline{48.0}  \\
        \bottomrule
        \end{tabular}
  \caption{
      \textbf{Generalization} reporting $\mAA(30)$ for 10 input frames.
      We first train on 41 CO3Dv2 seen categories.
      Testing is conducted on 11 unseen categories (top row), and on RealEstate10k (bottom)
      (\textbf{Bold} denotes the top result and an \underline{underline} signifies the second best).
    \label{tab:generalization_categories_co3dv2}
  }
}
\end{table}

\begin{figure}[t]
\centering\small%
\setlength\tabcolsep{0pt}%
\renewcommand{\arraystretch}{0.0}%
\newcommand{\nerffigwidth}{1.6cm}%
\begin{tabular}{ccccc}
\# frames & \method{RelPose}  & \method{COLMAP+SPSG} & \method{Ours} & Target \vspace{0.1cm} \\
\begin{minipage}[b][\nerffigwidth][c]{1cm} 10 \end{minipage} &
\includegraphics[trim=0mm 10mm 65mm 10mm, clip,width=\nerffigwidth]{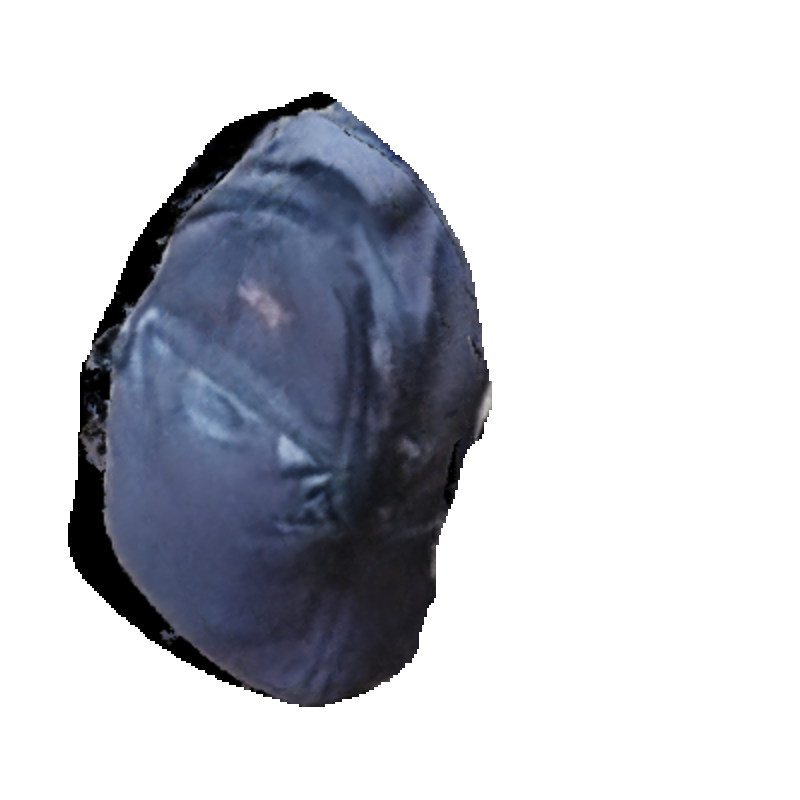} &
\includegraphics[trim=0mm 10mm 65mm 10mm,width=\nerffigwidth]{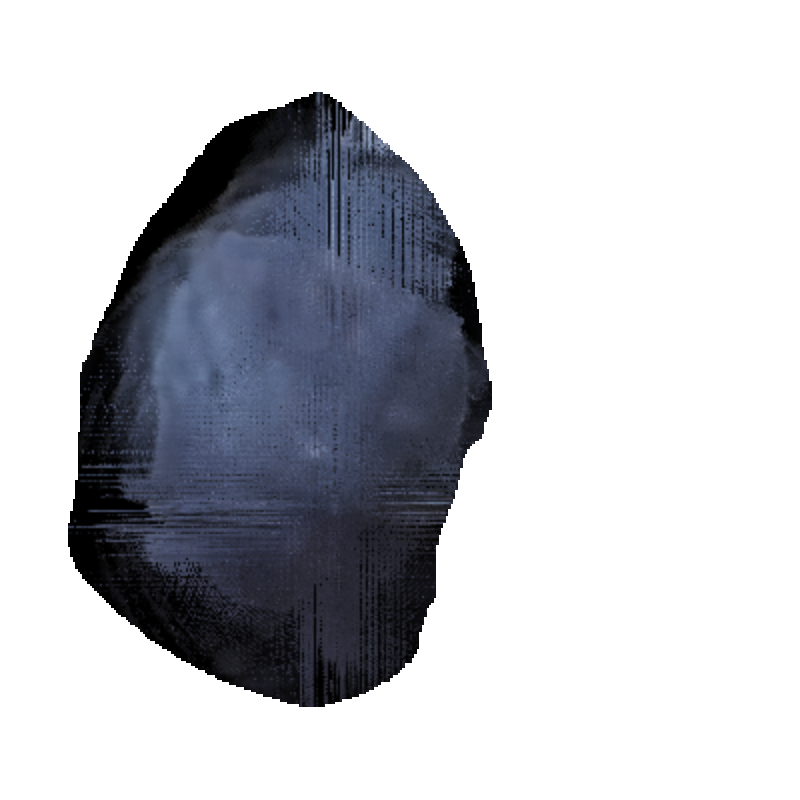} &
\includegraphics[trim=0mm 10mm 65mm 10mm,width=\nerffigwidth]{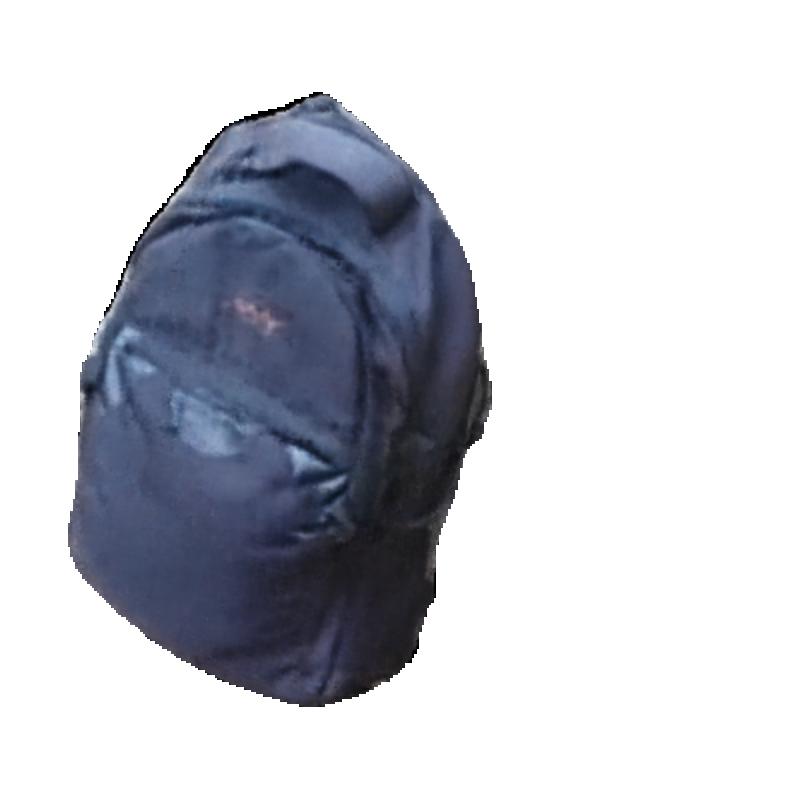} &
\includegraphics[trim=0mm 10mm 65mm 10mm,width=\nerffigwidth]{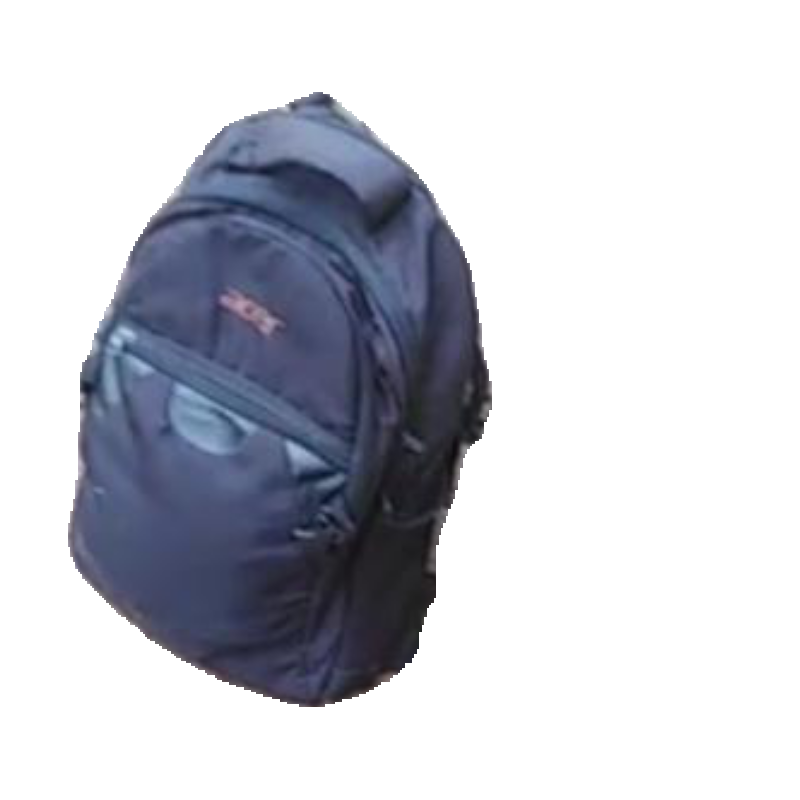} \\
\begin{minipage}[b][\nerffigwidth][c]{1cm} 20 \end{minipage} &
\includegraphics[trim=10mm 10mm 10mm 10mm,width=\nerffigwidth]{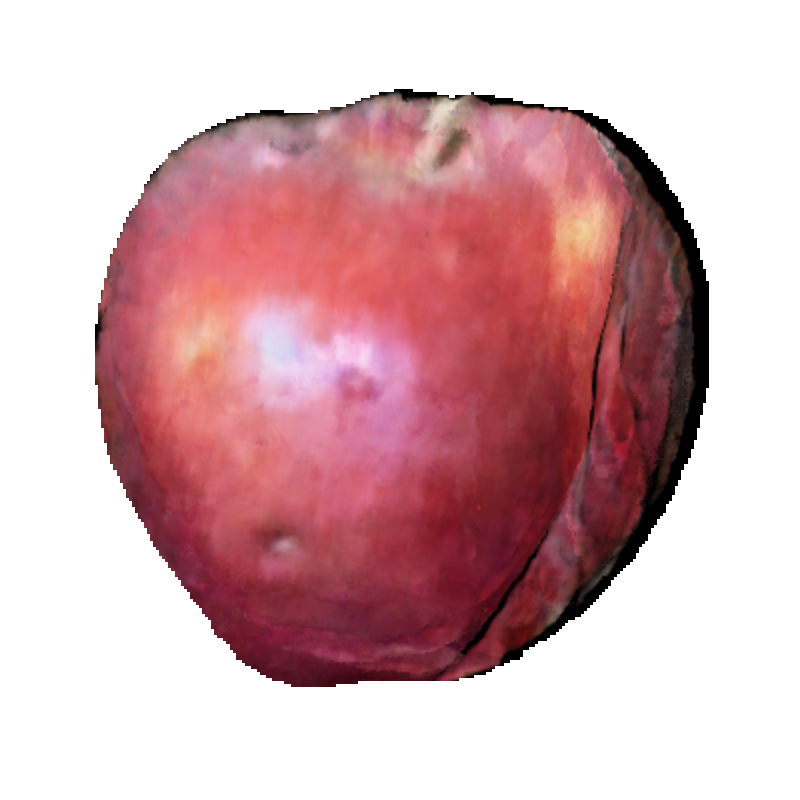} & \includegraphics[trim=10mm 10mm 10mm 10mm, width=\nerffigwidth]{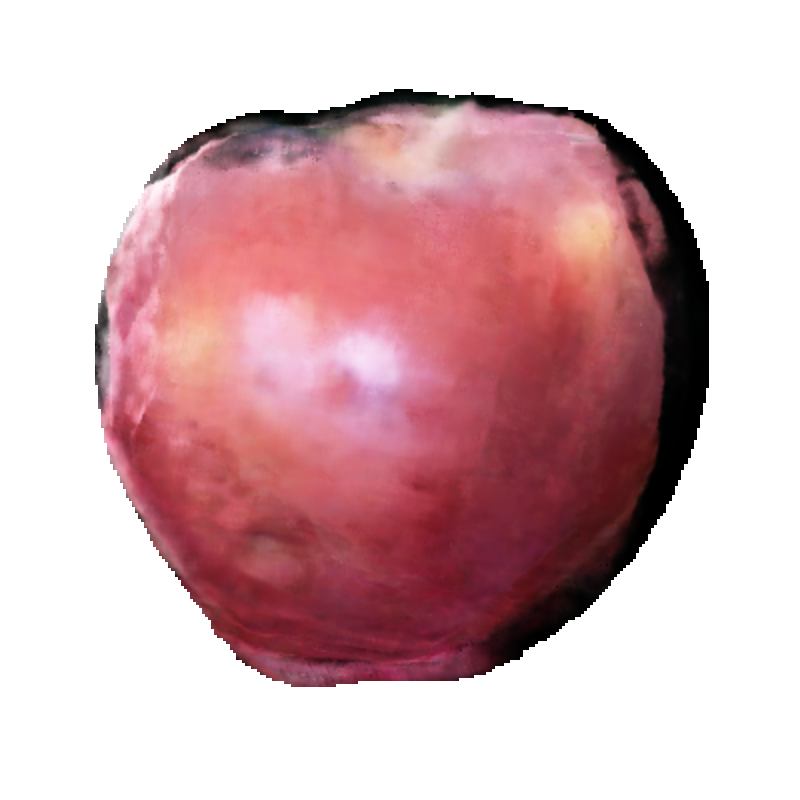} &
\includegraphics[trim=10mm 10mm 10mm 10mm, width=\nerffigwidth]{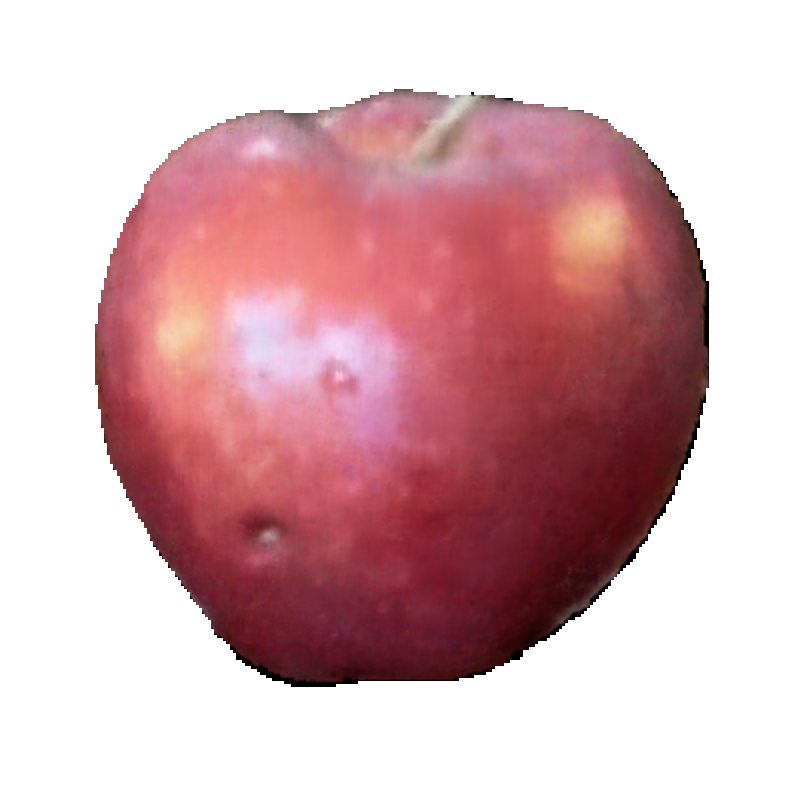} &
\includegraphics[trim=10mm 10mm 10mm 10mm, width=\nerffigwidth]{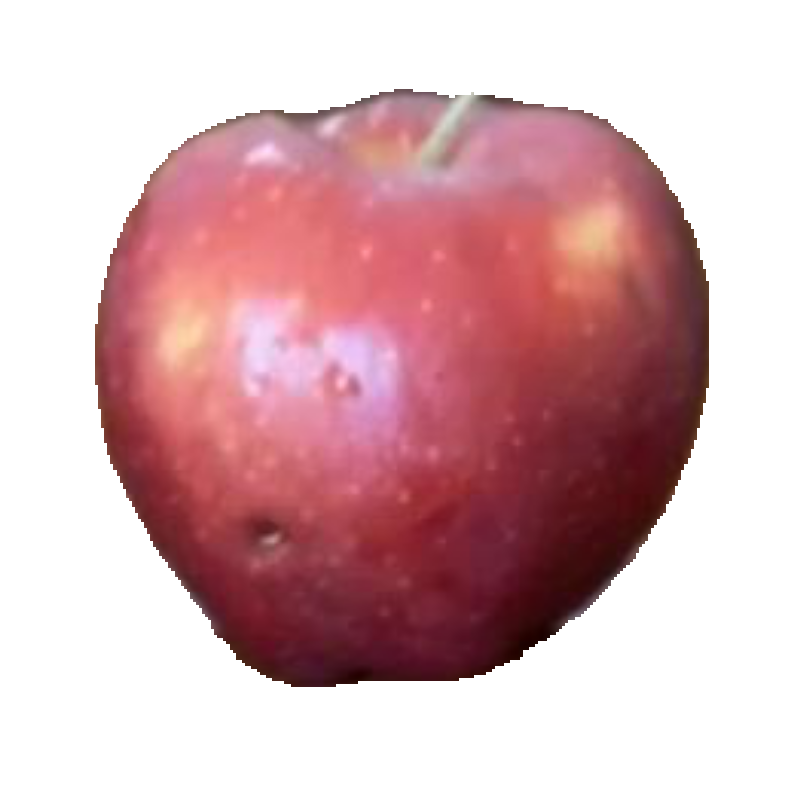} \\
\begin{minipage}[b][\nerffigwidth][c]{1cm} 50 \end{minipage} &
\includegraphics[trim=10mm 10mm 65mm 10mm,width=\nerffigwidth]{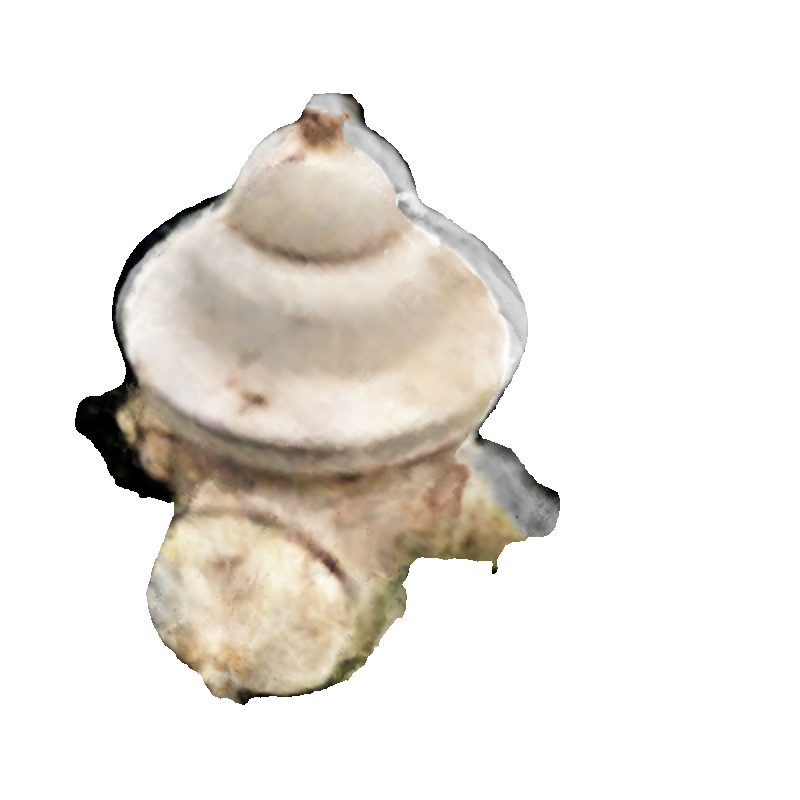} 
& \includegraphics[trim=10mm 10mm 65mm 10mm,width=\nerffigwidth]{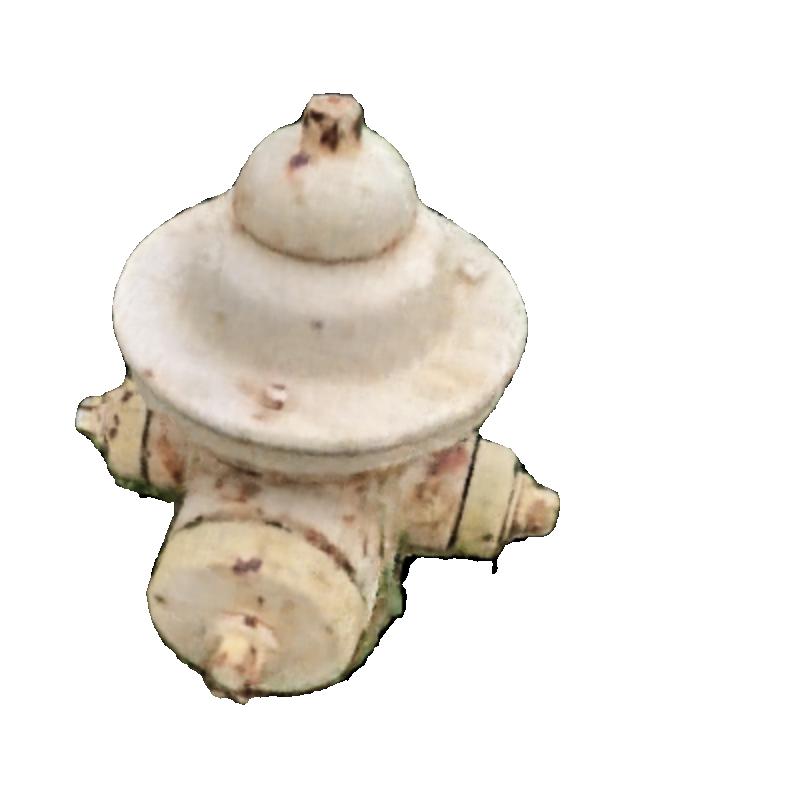} 
& \includegraphics[trim=10mm 10mm 65mm 10mm,width=\nerffigwidth]{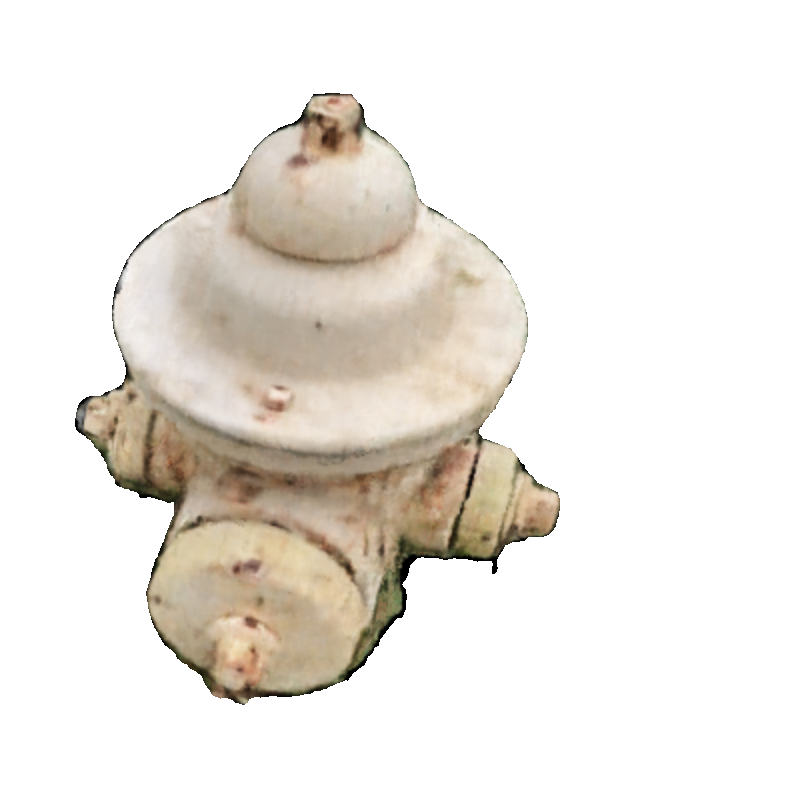} 
& \includegraphics[trim=10mm 10mm 65mm 10mm,width=\nerffigwidth]{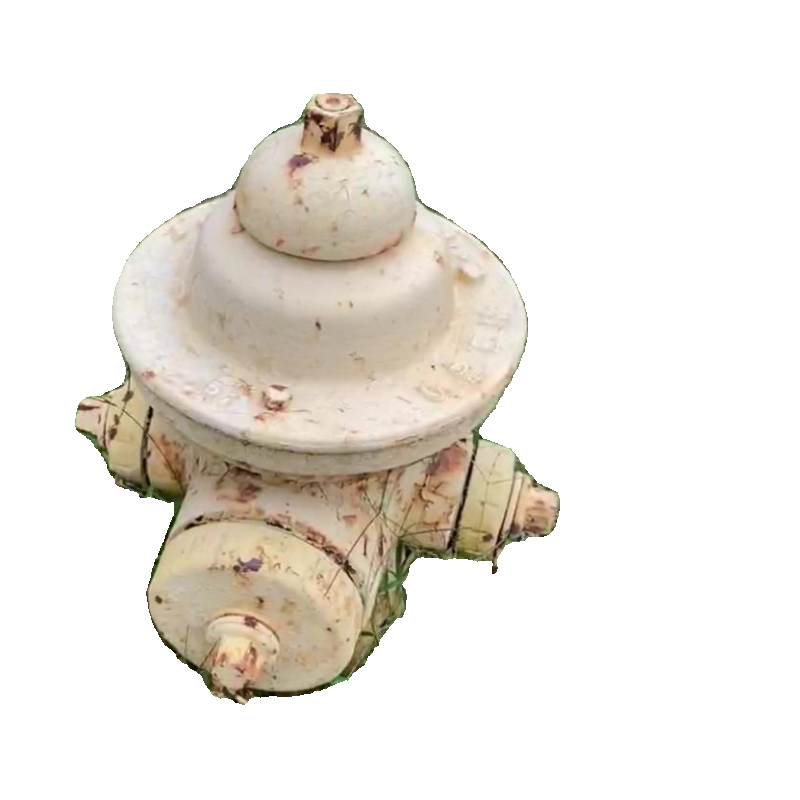} \\
\end{tabular}

\caption{
    \textbf{Synthesized novel views.} NeRF trained with camera poses estimated by various methods. This metric is more fair as it does not rely on GT pose annotations by another method.
    \label{fig:nvs_qual}
}
\end{figure}

\subsection{Novel-view synthesis.}
To evaluate the quality of the camera pose prediction for downstream tasks, we train NeRF models using predicted camera parameters and measure the RGB reconstruction error in novel views.
Note that, as opposed to the camera pose evaluation on CO3Dv2, here, we fairly evaluate against unbiased image ground truth.
We generate a dataset of 10, 20, and 50 frames for 50 random sequences of CO3Dv2.
Each sequence contains 4 validation frames with the remaining ones used to train the NeRF.
We report PSNR averaged over all validation frames as an indirect measure of camera pose accuracy. 
Furthermore, the experiment also evaluates the accuracy of the predicted intrinsics (focal lengths) since these are an inherent part of the NeRF's camera model significantly affecting the rendering quality. 

In \cref{tab:nerf_co3dv2}, our method outperforms \method{COLMAP+SPSG}, demonstrating the better suitability of our predicted cameras for NVS.
Moreover, \method{Ours + GT Focal Length}, which replaces the predicted focal lengths with the ground truth, is perfectly on par with \method{Ours}, signifying the reliability of our intrinsics.
\cref{fig:nvs_qual} provides the qualitative comparison.

\begin{table}[t]
    \centering\footnotesize
    \resizebox{0.8\linewidth}{!}{%
    \begin{tabular}{c|ccc}
        \toprule
        \multirow{2}{*}{Method} & \multicolumn{3}{c}{\# frames} \\
        & 10 & 20 & 50 \\ \midrule
        \method{RelPose} \cite{zhang2022relpose}$^\star$& 21.33  &23.12 &  25.09  \\
        \method{Ours} + GT Focal Length & 24.72 & 26.58 & 28.61 \\
        \midrule
        \method{COLMAP+SPSG} & 15.78 & 25.17 & \textbf{28.66} \\
        \method{Ours} & \textbf{24.37} &\textbf{26.96} & 28.53 \\        
        \bottomrule
    \end{tabular}}
  \caption{
      \textbf{Novel View Synthesis.} PSNR for NeRFs \cite{mildenhall_nerf_2020} trained on CO3Dv2 using cameras estimated by various methods.
    \method{RelPose}$^\star$ does not predict translation vectors and focal lengths, and uses the ground truth here instead. 
      \label{tab:nerf_co3dv2}
  }
  \vspace{-0.2cm}
\end{table}

\paragraph{Execution time.} Our method without GGS typically takes around 1 second for inference on a sequence of 20 frames, and enabling GGS increases the execution time to 60-90 seconds. 
GGS is currently unoptimized (a simple \textit{for} loop in Python),
compared to common C++ implementations for SfM methods which can be adopted here.

\section{Conclusion}
This paper presents \clearname, a learned camera estimator enjoying both the power of traditional epipolar geometry constraint and diffusion model. 
We show how the diffusion framework is ideally compatible with the task of camera parameter estimation. 
The iterative nature of this classical task is mirrored in the denoising diffusion formulation. 
Additionally, point-matching constraints between image pairs can be used to guide the model and refine the final prediction.
In our experiments, we improve over traditional SfM methods such as {COLMAP}, as well as the learned approaches.
We are able to show improvements regarding the pose  prediction accuracy as well as on the novel-view synthesis task, which is one of the most popular current applications of {COLMAP}.
Finally, we are able to demonstrate that our method can overcome one of the main limitations of learned methods:  generalization across datasets, even when trained on a dataset with different pose distributions.

\vspace{5mm}
\subsection*{Acknowledgements.}
We would like to thank Nikita Karaev, Luke Melas-Kyriazi, and Shangzhe Wu for their insightful discussions.
We appreciate the great help from Jason Y. Zhang for sharing the baseline/benchmark code.
Jianyuan Wang is supported by Facebook Research. Christian Rupprecht is supported by ERC-CoG UNION 101001212 and VisualAI EP/T028572/1.


\appendix


\begin{figure*}[h] 
\centering
\includegraphics[width=\textwidth]{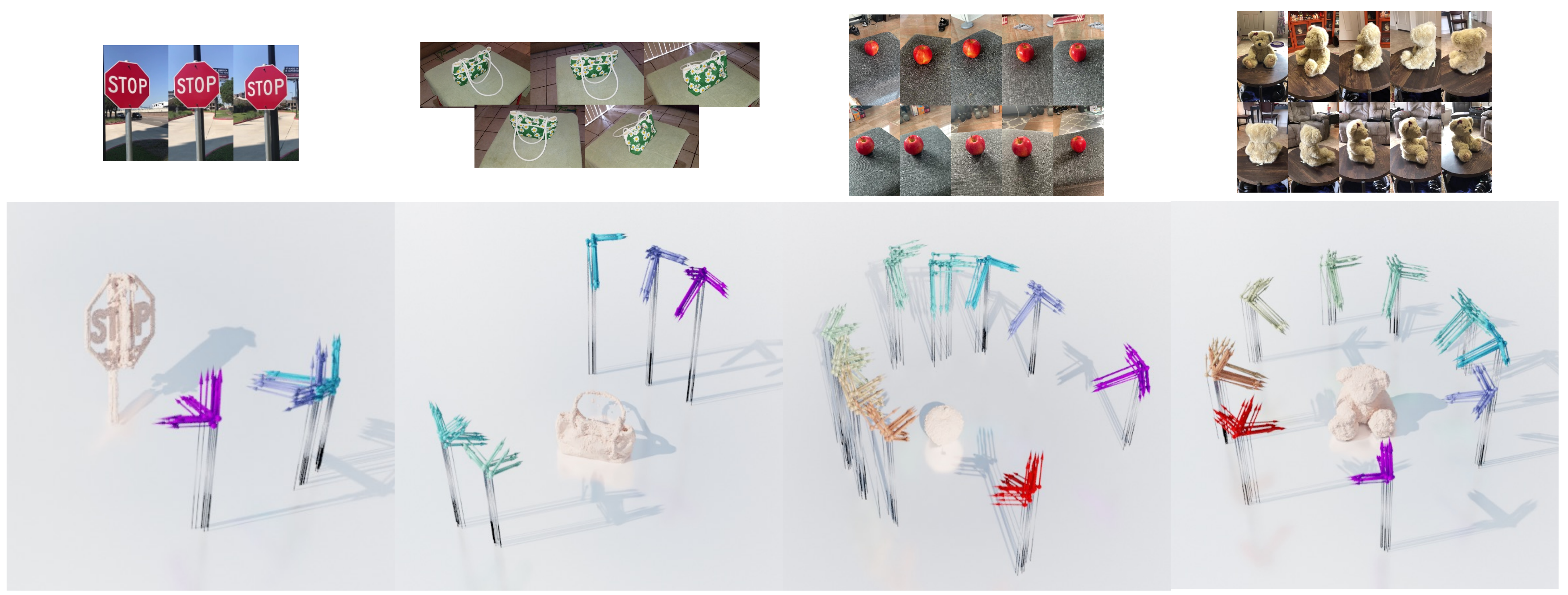}
\caption{%
\textbf{Pose uncertainty} visualizing multiple samples from $p(x | \I)$ conditioned on the same set of input images $\I$. The cameras predicted for the same frame are indicated with identical colors. 
} \label{fig:pose_confidence}%
\end{figure*}


\section{Implementation Details}

\begin{figure*}[h]
  \centering
  \includegraphics[width=0.9\linewidth]{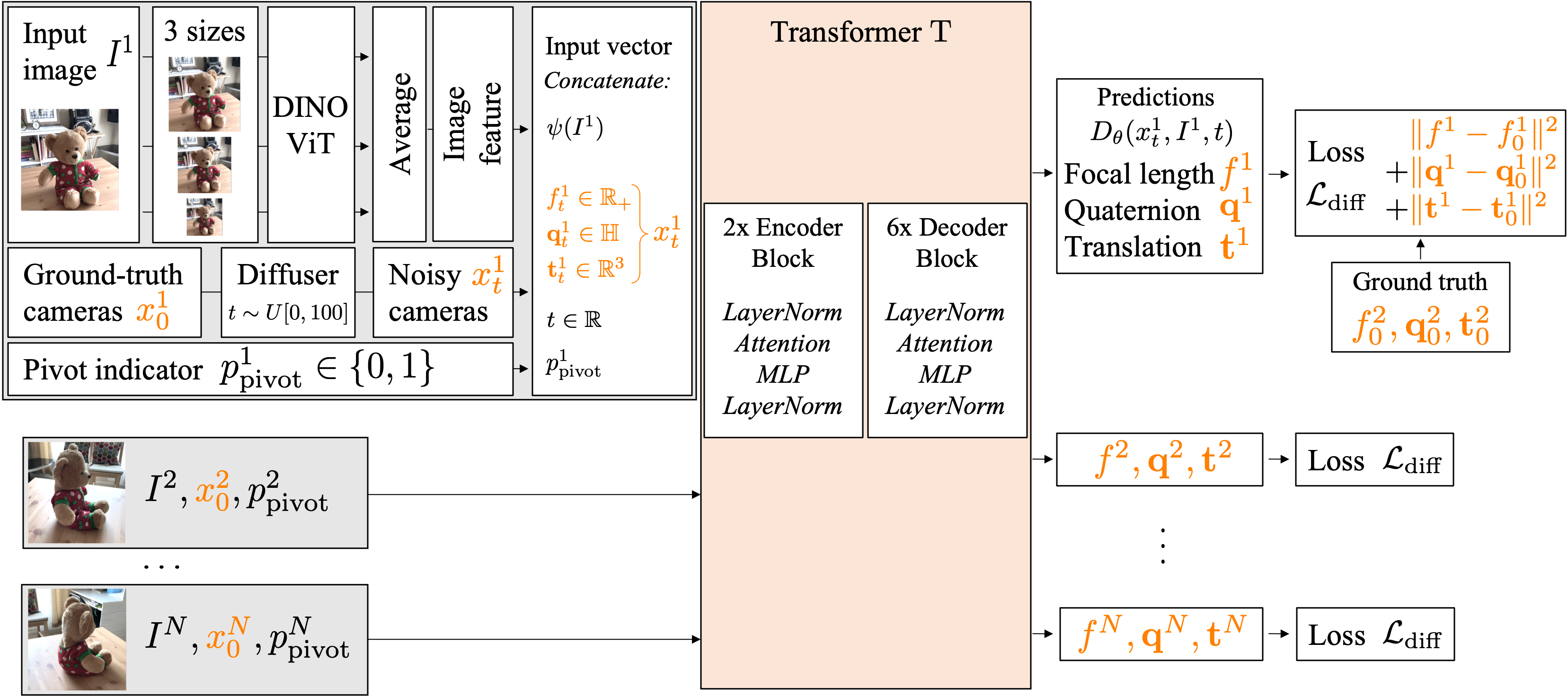}
  \caption{
  \textbf{Overview of our architecture} depicting a single training-mode forward pass.
  }
 \label{fig:suppl_overview}
\end{figure*}

In this section, we provide more method details.
Additionally, \cref{fig:suppl_overview} illustrates a single training-mode forward pass of \clearname.

\paragraph{Feature Extraction.} We use the pretrained DINO ViT-S16 model~\cite{caron2021dino} as our feature extraction backbone. The model and the weight are available in its \href{https://github.com/facebookresearch/dino}{public repository}. We first center-crop the input images and resize them to a resolution of 224$\times$224. Similar to~\cite{caron2021dino}, we then respectively resize the images to $1$, $\frac{1}{2}$, and $\frac{1}{3}$ of the input resolution (224), and average their features to achieve a multi-scale understanding. The weights of the DINO model are optimized during our training.

\paragraph{Representation and Canonicalization.} 
We represent the camera poses with $\left( (\hat{f}^i, \q^i, \T^i) \right)_{i=1}^N$.
This representation has a dimensionality of 8: 1 for focal length $\hat{f}$, 4 for quaternion rotation $\q$, and 3 for translation $\T$. 
As mentioned in the main paper, for each sequence, we randomly chose one input frame as the `canonical' (pivot) one. 
Specifically, we reorient the coordinate system of the sequence so that it is centered at the pivot camera. 
This transformation results in the pivot camera being positioned at the origin with no translation, and with an identity rotation matrix. 
We explicitly provide this information to the network by utilizing a one-hot pivot flag. 
Furthermore, in order to prevent overfitting to scene-specific translation scales, we  normalize the translation vectors by the median norm. 

More specifically, given a batch of scene-specific training SfM extrinsics
$\{\hat{g}^1, ... \hat{g}^N\} = \mathcal{T}_j \in \mathcal{T}$,
the transformer $T$ ingests normalized extrinsics $g^i = s((\hat{g}^\star)^{-1} \hat{g}^i)$ which are expressed relative to a randomly selected pivot camera $\hat{g}^\star \in \mathcal{T}_j$, where $s(\cdot)$ denotes scale normalization which divides the translation component $\T$ of the input $\mathbb{SE}(3)$ transformation by the median of the norms of the pivot-normalized translations. Focal lengths and principal points remain unchanged in the whole process. 
To avoid the extreme cases brought by canonicalization of outliers, we clamp the input and estimated translation vectors at a maximum absolute value of 100. 
We also clamp the predicted focal lengths by a maximum value of 20.

\begin{figure*}[h]
  \centering
  \input{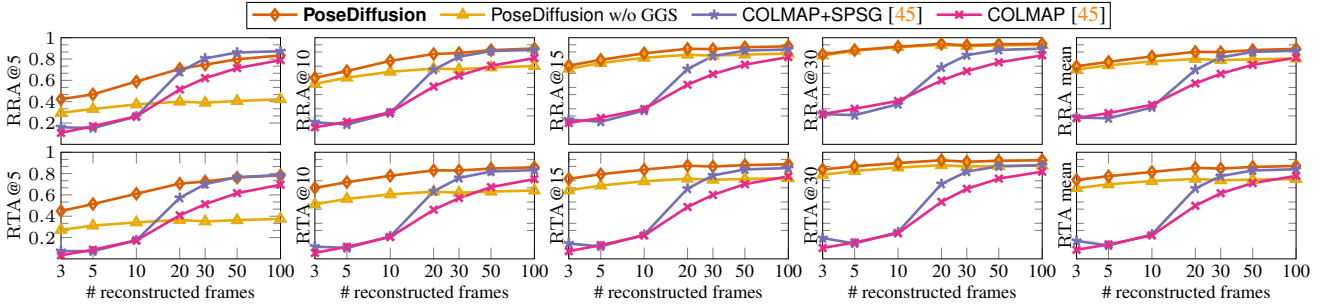}
\newcommand{\metricname}[1]{%
    \IfEqCase{#1}{
        {rre}{{$\RRE$}}
        {rte}{{$\RTE$}}
        {are}{{$\ARE$}}
        {ate}{{$\ATE$}}
    }[{}]%
}
\newcommand{\accfigureplots}[3]{
    \addplot [color=ourswoggs,mark=triangle,line width=1pt] table [col sep=space,x=numframe,y=ourswoggs] {#3#2#1.csv};{\label{plots:co3d:ourswoggs}}
    \addplot [color=ours,mark=diamond,line width=1pt] table [col sep=space,x=numframe,y=ours] {#3#2#1.csv};
    {\label{plots:co3d:ours}}
    \addplot [color=colmapspsg,mark=star,line width=1pt] table [col sep=space,x=numframe,y=colmapspsg] {#3#2#1.csv};{\label{plots:co3d:colmapspsg}}
    \addplot [color=colmapsift,mark=x,line width=1pt] table [col sep=space,x=numframe,y=colmapsift] {#3#2#1.csv};{\label{plots:co3d:colmapsift}}  
}
\newcommand{\accfigure}[5]{%
    \nextgroupplot[
        title={},
        xlabel={
            \IfEqCase{#2}{{rte}{{\# reconstructed frames}}}[{}]
        },
        ylabel={
            \metricname{#2}%
            \IfEqCase{#1}{{mean}{{ mean}}}[{@#1}]
        },
        font=\scriptsize, 
        width=4.5cm,
        height=3.0cm,
        line width=0.25pt,
        mark size=1pt,
        xmajorticks={#4}, 
        xtick={3,5,10,20,30,50,100},
        xticklabels={3,5,10,20,30,50,100},
        ymajorticks={#5},
        ytick = {0.2,0.4,0.6,0.8,1.0},
        ymin=0,
        ymax=1.0,
        xmin=3,
        xmax=100,
        minor tick num=2,
        inner sep=1pt,
        outer sep=1pt,
        mark size=2pt,
        xmode=log,
    ]
    \accfigureplots{#1}{#2}{#3}
}
\centering%
\begin{tikzpicture}
\begin{groupplot}[
    group style={group size=5 by 2,vertical sep={3pt}, horizontal sep={13pt}},
    legend style={legend columns=4},
    legend entries={\namewoggs,\name,COLMAP+SPSG},
    legend cell align=left,
    legend to name=grouplegend,
    legend style={draw=none, fill opacity=0.5, text opacity = 1,row sep=-1pt,font=\scriptsize},
]
    \accfigure{5}{rre}{co3dv2}{false}{true}
    \coordinate (top) at (rel axis cs:0,1);
    \accfigure{10}{rre}{co3dv2}{false}{false}
    \accfigure{15}{rre}{co3dv2}{false}{false}
    \accfigure{30}{rre}{co3dv2}{false}{false}
    \accfigure{mean}{rre}{co3dv2}{false}{false}
    \accfigure{5}{rte}{co3dv2}{true}{true}
    \accfigure{10}{rte}{co3dv2}{true}{false}
    \accfigure{15}{rte}{co3dv2}{true}{false}
    \accfigure{30}{rte}{co3dv2}{true}{false}
    \accfigure{mean}{rte}{co3dv2}{true}{false}
    \coordinate (bot) at (rel axis cs:1,0);
\end{groupplot}
\path (top|-current bounding box.north)--
      coordinate(legendpos)
      (bot|-current bounding box.north);
\footnotesize%
\matrix[%
    matrix of nodes,
    anchor=south,
    draw,
    inner sep=0.1em,
    draw
]at([yshift=0ex]legendpos)
{
    \ref{plots:co3d:ours} & \textbf{\name} & [3pt]
    \ref{plots:co3d:ourswoggs} & \namewoggs & [3pt]
    \ref{plots:co3d:colmapspsg} & \method{COLMAP+SPSG} \cite{schonberger_structure--motion_2016} & [3pt]
    \ref{plots:co3d:colmapsift} & \method{COLMAP} \cite{schonberger_structure--motion_2016} \\ 
};
\end{tikzpicture}%
  \caption{
  \textbf{Camera accuracy on CO3Dv2} with a larger number of input views (up to 100).
  We compare to {COLMAP} / {COLMAP+SPSG} and omit comparison to {RelPose} because it is prohibitively memory-demanding for a larger number of frames. Our method is competitive even with 100 frames.
  }
 \label{fig:co3dv2_suppl_plots}
\end{figure*}

\paragraph{Architecture.} 
For the input of the denoiser $\mathcal{D}_\theta$, we concatenate the input poses $x_t^i$, the diffusion time $t$, and the feature embeddings $\psi(I^i)$ of the input images $I^i$. 
Specifically, we first project the concatenated input poses $x_t^i \in \R^8$ and steps $t \in \R$ to a feature vector with 96 dimensions (dim) by a linear transformation. 
Next, we concatenate the 96$\text{-}$dim feature vector with $x_t^i$, $t$, and 385$\text{-}$dimensional image features $\psi(I^i)$, fed into the denoiser.
The image feature embedding $\psi(I^i)$ comprises 384$\text{-}$dim DINO features and the one-dimensional binary pivot camera flag $p^i_\text{pivot} \in \{0, 1\}$.

The denoiser $\mathcal{D}_\theta$ adopts a classic Transformer architecture.
We use the \href{https://pytorch.org/docs/stable/generated/torch.nn.Transformer.html}{built-in} implementation of PyTorch.   
%
Our denoiser has 8 encoder layers and does not use decoder layers. 
The number of heads is set to 4, and the dimension of the feedforward network is 1024.
%
%
The output features of the transformer are passed into a two-layer MLP to give the final prediction.
The hidden dimension of the final MLP is 128 and the output dimension is 8.


\paragraph{Diffusion Model.} We use the \href{https://github.com/lucidrains/denoising-diffusion-pytorch}{PyTorch implementation} of DDPM~\cite{ho_denoising_2020}. We set the total number of diffusion sampling steps $\mathrm{T}$ as $100$. Following the default setting of DDPM, the forward process variance ($\beta_t$) increases linearly from $10^{-3}$ to $0.2$. 
We empirically chose the ``$x_0$ formulation'' of DDPM because it exhibits a more stable training and marginally better performance than predicting the noise profile.
The other hyperparameters were kept at their default values as per the utilized DDPM codebase.




\paragraph{GGS.} 
The  guidance strength $s$ is set adaptively to
$s = \min(\alpha \frac{\| \mu_t \| }{ \| \nabla p(\I | x) \|}, 1.0)$, with $\alpha=0.0001$ to stabilize training.
We skip the GGS process if no matches were discovered between any pair of input frames.


\paragraph{Training.} We train our model on 8 NVIDIA Tesla V100 GPUs, each with 192 images. For each sequence, we randomly conduct \href{https://pytorch.org/vision/main/generated/torchvision.transforms.ColorJitter.html}{color-jitter} augmentation to all images in each batch. Additionally, with a probability of 0.15, we randomly turn each training image to its gray-scale form.
To ensure stable training, we rescale the optimization gradient so that its norm does not exceed $1.0$. The whole training pipeline is implemented using \href{https://github.com/facebookresearch/pytorch3d}{PyTorch3D}.

\paragraph{Evaluation.}  
As mentioned, we align the predicted camera poses to the ground truth ones by a single optimal similarity before evaluation, which is implemented by Umeyama's algorithm~\cite{umeyama1991least}. 
The latter aligns the 3D locations of the optical centers of the predicted cameras to the centers of the corresponding ground truth cameras.


\paragraph{NeRF.} The training and evaluation of our NeRF experiments leverage the  \href{https://github.com/facebookresearch/pytorch3d/tree/main/projects/implicitron_trainer}{Implicitron} framework.
Each NeRF model was trained using the default parameters of the framework.
We empirically verified that using a single focal length comprising the average over all frame-specific focal length predictions provides better performance.
To ensure reconstructibility of the evaluation sequences, we first train NeRF with ground-truth camera poses and, select only the ones where training/evaluation with 8/2 views gives PSNR of 25 or better.



\paragraph{Fundamental Matrix Derivation.}
Epipolar geometry, \ie the relationship between points and lines of two cameras observing the same scene, can be algebraically represented via the fundamental matrix $F \in \R^{3 \times 3}$.
In more detail, denote $(x^i, x^j)$ the parameters of the camera pair, where $x = (K, g)$ consists of intrinsics
$K \subset \R^{3 \times 3}$
and extrinsics
$g \in \mathbb{SE}(3)$.
The extrinsics $g$ can be further expressed as a rotation matrix and the translation vector  $(R \in \mathbb{SO}(3), \T \in \R^3)$.
Using the latter, we define a $3 \times 4$ projection matrix $M = K \ [R\ | \ \T ]$.
%


Assume a point ${\tilde{\p}}$ in the camera plane defined by $M^i$.
The ray back-projected from ${\tilde{\p}}$ by $M^i$ can be written as $[M^{i}]^{+} {\tilde{\p}}+\lambda C$,
where $[M^{i}]^{+}$ is the pseudo-inverse of $M^i$, and $C$ is the camera center so that $M^i C = \mathbf{0}$.
The scalar $\lambda \in \R$ parametrizes the ray. 
Setting $\lambda=0$ and $\lambda=\infty$ yields $[M^{i}]^{+} {\tilde{\p}}$ and the camera center $C$ respectively.
These two points will be imaged at the second image plane $M^j$ as $M^j [M^{i}]^{+} {\tilde{\p}}$ and $M^j C$.
The epipolar line $l^j$ is defined as the line connecting these two points, \ie, $l^j = (M^j C) \times M^j [M^{i}]^{+} {\tilde{\p}}$.
The fundamental matrix $F$ is defined as the mapping from a point in the first image plane to its corresponding epipolar line in the second plane, \ie $l^j = F {\tilde{\p}}$. 
Therefore, we obtain $F = (M^j C) \times M^j [M^{i}]^{+}$.
It is worth noting that the point ${\tilde{\p}}$ is removed from the formulation of $F$, because $F$ is the relationship between two image planes, and is constant for all the points in one image plane.
For more details, please refer to~\cite{hartley_multiple_2004}.





\section{Evaluation with More Frames}
In \cref{fig:co3dv2_suppl_plots}, we provide camera accuracy metrics on CO3Dv2 when more frames are reconstructed.
Even though, in the many-frame regime, the evaluation puts {COLMAP} to unfair advantage since the latter produced the ground-truth camera annotations, we note that \clearname{} performs on par with {COLMAP+SPSG} for all numbers of reconstructed frames.

\section{Ablation Studies and Analysis}
Unless otherwise stated, all ablation studies are conducted on CO3Dv2.

\paragraph{Camera Pose Uncertainty.}  One inherent advantage of utilizing the diffusion model for camera pose estimation is its probabilistic nature. It is well-known that few-view camera pose estimation is a non-deterministic problem, where multiple pose combinations may be all reasonable for a set of images. 
We provide a visualization in \cref{fig:pose_confidence} to verify that our method can provide several reasonable pose sets $x$ for the same input frames $\I$. 




\paragraph{Backbone.}
To explore the effect of upstream feature quality, we try different feature extraction backbones as shown in ~\cref{tab:ab_backbone}.
ResNet50 trained in a self-supervised manner (DINO ResNet50~\cite{caron2021dino}) performs better than ResNet50 trained by supervised image classification~\cite{he2016resnet}.
The DINO ViT model~\cite{caron2021dino} shows the best performance.

\begin{table}[t]
    \centering\footnotesize
        \resizebox{\linewidth}{!}{%
    \setlength\tabcolsep{3pt}%
    \begin{tabular}{c|ccc}
        Backbone & {ResNet50 (sup.~\cite{he2016resnet})}  & {ResNet50 (DINO~\cite{caron2021dino})}  & {ViT-S16 (DINO~\cite{caron2021dino})} \\
    \midrule
    mAA(30) & 63.1 & 64.3 & \textbf{66.5} \\
        \bottomrule
    \end{tabular}
    }
  \caption{
      \textbf{Performance of different feature backbones}. With other settings unchanged, we evaluate different feature extraction backbones on CO3Dv2.
      \label{tab:ab_backbone}
  }
\end{table}



\paragraph{Diffusion Steps.} Differently from the original application in image generation (which requires 1000 diffusion steps), the results in \cref{tab:ab_steps} show that a moderate number of sampling steps ($T=100$) suffice. Therefore, we use $T=100$ for all the experiments if not further specified.



\paragraph{Importance of Background.} We have observed that our method can produce favorable results even when the object is nearly symmetrical. One plausible explanation for this is that the model utilizes cues from the textured background to estimate relative poses (which is valid and desired in SfM). In order to test this hypothesis, we conducted an experiment where we mask the background. 
In \cref{tab:ab_background} the performance of the model declines significantly when we replace the background pixels with black, which supports our intuition.




\begin{table}[t]
    \centering\small
    \setlength\tabcolsep{3pt}%
    \begin{tabular}{c|cccc}
        \toprule
        \# diffusion steps $T$ & 30  & 50  & 100 & 500 \\
    \midrule
    mAA(30) &  62.5 & 66.1  & \textbf{66.5} & 65.3 \\ 
        \bottomrule
    \end{tabular}
  \caption{
      \textbf{The effect of the number of sampling steps $T$}.
      We evaluate  the value of the diffusion sampling steps $\mathrm{T}$ from 30 to 500.
      \label{tab:ab_steps}
  }
\end{table}



\begin{table}[t]
    \centering\small
    \setlength\tabcolsep{3pt}%
    \begin{tabular}{c|cc}
        \toprule
         & {w/o background}  & {w background} \\
    \midrule
    mAA(30) & 57.0 & \textbf{66.5} \\
        \bottomrule
    \end{tabular}
  \caption{
      \textbf{Effect of background pixels on CO3Dv2}.
      We compare camera accuracy attained when letting {\clearname} observe background pixels (w background) and when using the foreground masks to mask-out the background (w/o background).
      \label{tab:ab_background}
  }
\end{table}







\section{Future Work}
%
%
Looking ahead, we plan to extend the current framework to a self-supervised manner, which would eliminate the need for high-quality ground truth camera poses. This would enable the model to take advantage of numerous Internet data and expand its applicability to a wider range of data distributions. 
%
Additionally, our method can serve as a robust initialization for classic Bundle Adjustment frameworks like COLMAP, which could further enhance the accuracy of the pose estimates without the need for the costly and complex iterative SfM process.

{\small
\bibliographystyle{ieee_fullname}
\bibliography{refsmore,largelibrary}
}


\end{document}